\documentclass[a4paper,twoside]{article}

\usepackage{epsfig}
\usepackage{subcaption}
\usepackage{calc}
\usepackage{amssymb}
\usepackage{amstext}
\usepackage{amsmath}
\usepackage{amsthm}
\usepackage{multicol}
\usepackage{pslatex}
\usepackage{apalike}
\usepackage{algorithm2e}
\usepackage[bottom]{footmisc}
\usepackage{url}
\usepackage{placeins}  
\usepackage{float}

\usepackage{bm}
\usepackage[table]{xcolor} 
\usepackage{booktabs}
\usepackage{graphicx}
\usepackage{pgfplots}

\pgfplotsset{compat=1.18}
\usetikzlibrary{shapes,plotmarks}
\graphicspath{{./figures/}}

\definecolor{rowgray}{gray}{0.95}
\definecolor{lightyellow}{rgb}{1.0, 1.0, 0.8}
\definecolor{lightorange}{rgb}{1.0, 0.9, 0.8}
\definecolor{lightred}{rgb}{1.0, 0.85, 0.85}

\usepackage{SCITEPRESS}     

\begin{document}

\title{Fast-HaMeR: Boosting Hand Mesh Reconstruction using Knowledge Distillation}

\author{\authorname{
Hunain Ahmed Jillani\sup{1}\orcidAuthor{0009-0000-9286-898X}, 
Ahmed Tawfik Aboukhadra\sup{1,2}\orcidAuthor{0009-0005-6111-1366}, 
Ahmed Elhayek\sup{3}\orcidAuthor{0000-0002-5919-7202}, 
Jameel Malik\sup{4}\orcidAuthor{0000-0002-7528-7240}, 
Nadia Robertini\sup{2}\orcidAuthor{0009-0007-2861-3326}, 
Didier Stricker\sup{1, 2}\orcidAuthor{0000-0002-5708-6023}}
\affiliation{\sup{1}RPTU, Kaiserslautern, Germany;
\sup{2}DFKI-AV, Kaiserslautern, Germany;
\sup{3}UPM, Saudi Arabia;
\sup{4}NUST-SEECS, Islamabad, Pakistan;
}
}



\abstract{
Fast and accurate 3D hand reconstruction is essential for real-time applications in VR/AR, human-computer interaction, robotics, and healthcare. 
Most state-of-the-art methods rely on heavy models, limiting their use on resource-constrained devices like headsets, smartphones, and embedded systems.
In this paper, we investigate how the use of lightweight neural networks, combined with Knowledge Distillation, can accelerate complex 3D hand reconstruction models by making them faster and lighter, while maintaining comparable reconstruction accuracy.
While our approach is suited for various hand reconstruction frameworks, we focus primarily on boosting the HaMeR model, currently the leading method in terms of reconstruction accuracy. 
We replace its original ViT-H backbone with lighter alternatives, including MobileNet, MobileViT, ConvNeXt, and ResNet, and evaluate three knowledge distillation strategies: output-level, feature-level, and a hybrid of both.
Our experiments show that using lightweight backbones that are only $35\%$ the size of the original achieves $1.5x$ faster inference speed while preserving similar performance quality with only a minimal accuracy difference of $0.4mm$.    
More specifically, we show how output-level distillation notably improves student performance, while feature-level distillation proves more effective for higher-capacity students.
Overall, the findings pave the way for efficient real-world applications on low-power devices.
The code and models are publicly available under \url{https://github.com/hunainahmedj/Fast-HaMeR}.
}

\onecolumn \maketitle \normalsize \setcounter{footnote}{0} \vfill

\section{INTRODUCTION}

\setlength\fboxsep{0pt} 
\setlength{\fboxrule}{0.5pt}

\begin{figure}[t!]
  \centering
  \begin{subfigure}[t]{0.11\textwidth}
    \fcolorbox{black}{white}{\includegraphics[width=\linewidth]{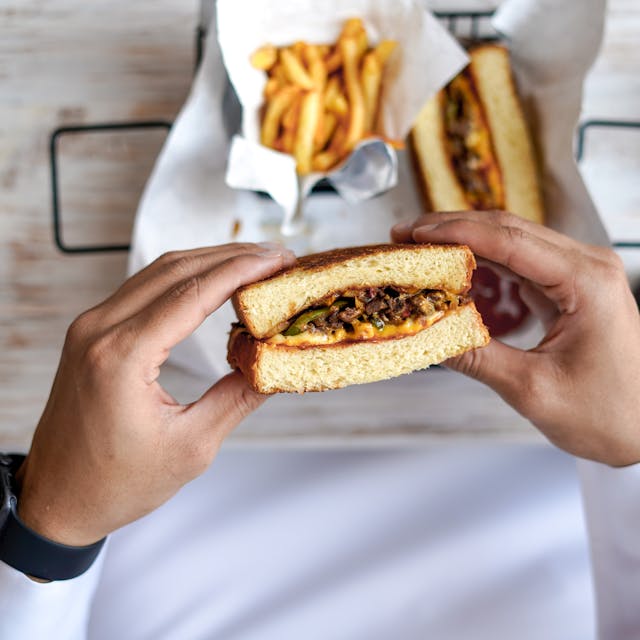}}
  \end{subfigure}
  \begin{subfigure}[t]{0.11\textwidth}
    \fcolorbox{black}{white}{\includegraphics[width=\linewidth]{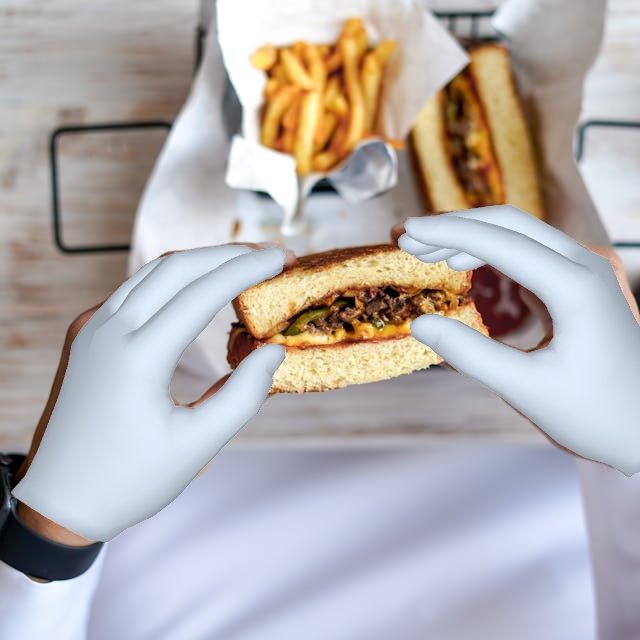}}
  \end{subfigure}
  \begin{subfigure}[t]{0.11\textwidth}
    \fcolorbox{black}{white}{\includegraphics[width=\linewidth]{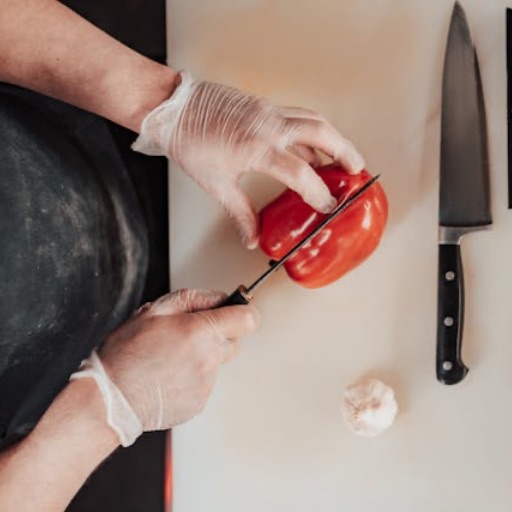}}
  \end{subfigure}
  \begin{subfigure}[t]{0.11\textwidth}
    \fcolorbox{black}{white}{\includegraphics[width=\linewidth]{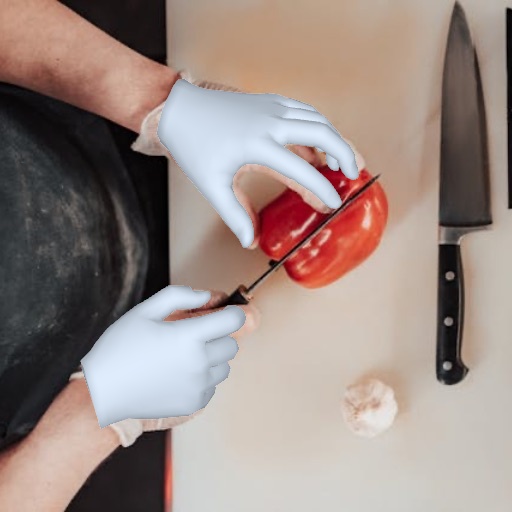}}
  \end{subfigure}
  \begin{subfigure}[t]{0.11\textwidth}
    \fcolorbox{black}{white}{\includegraphics[width=\linewidth]{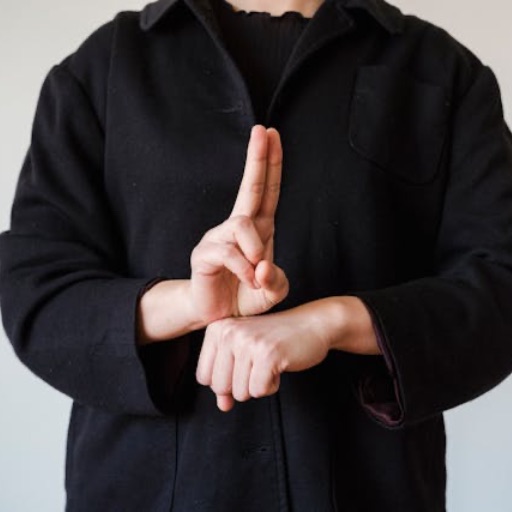}}
  \end{subfigure}
  \begin{subfigure}[t]{0.11\textwidth}
    \fcolorbox{black}{white}{\includegraphics[width=\linewidth]{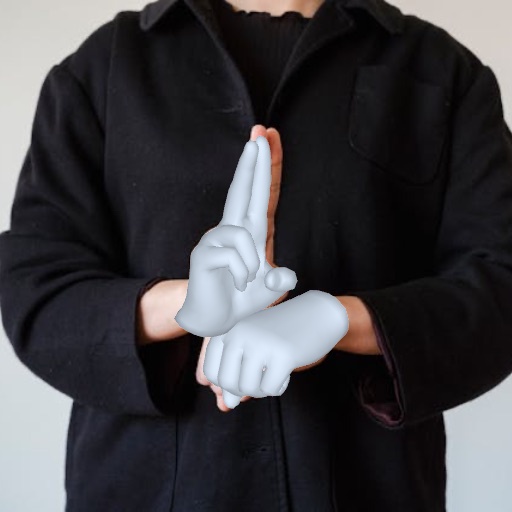}}
  \end{subfigure}
  \begin{subfigure}[t]{0.11\textwidth}
    \fcolorbox{black}{white}{\includegraphics[width=\linewidth]{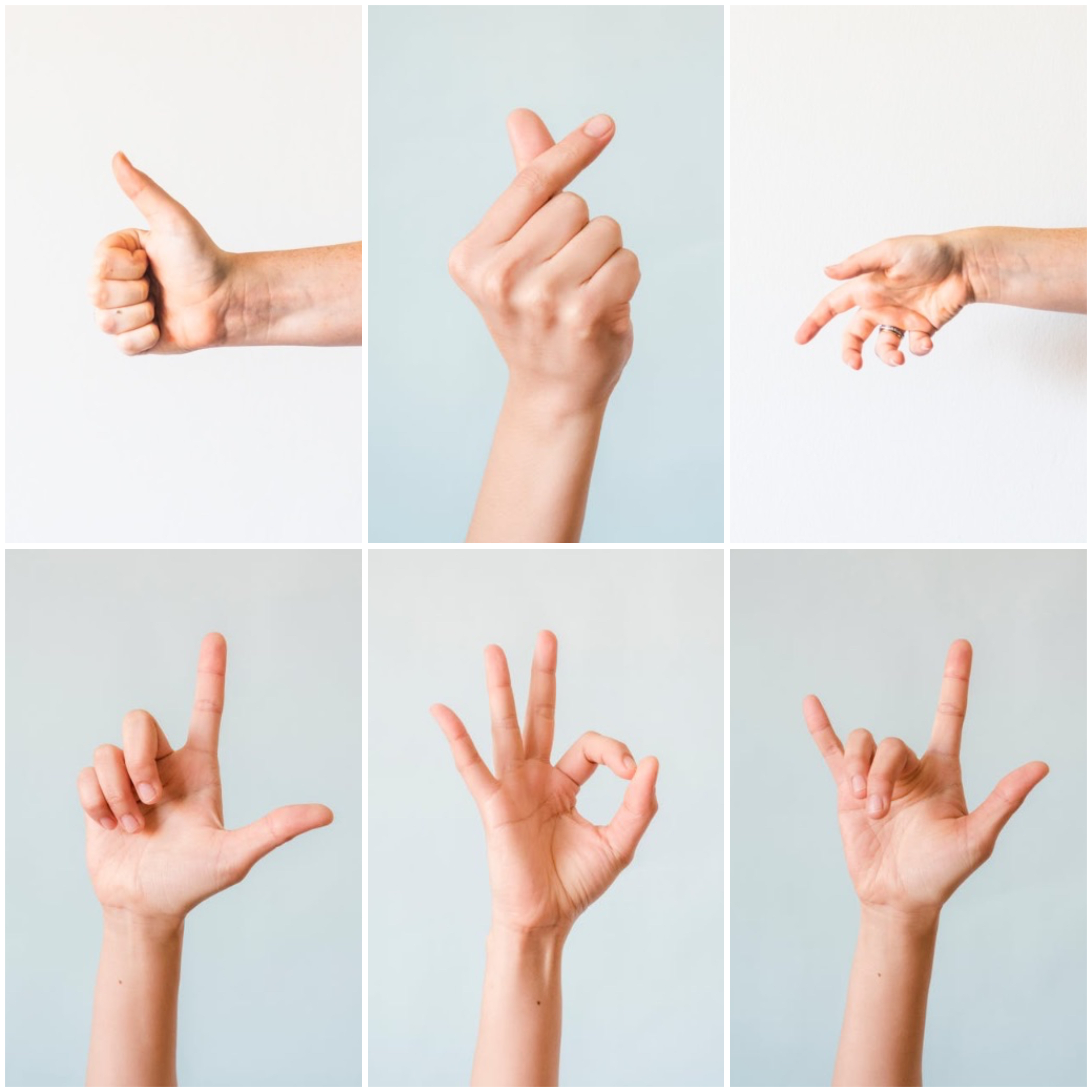}}
  \end{subfigure}
  \begin{subfigure}[t]{0.11\textwidth}
    \fcolorbox{black}{white}{\includegraphics[width=\linewidth]{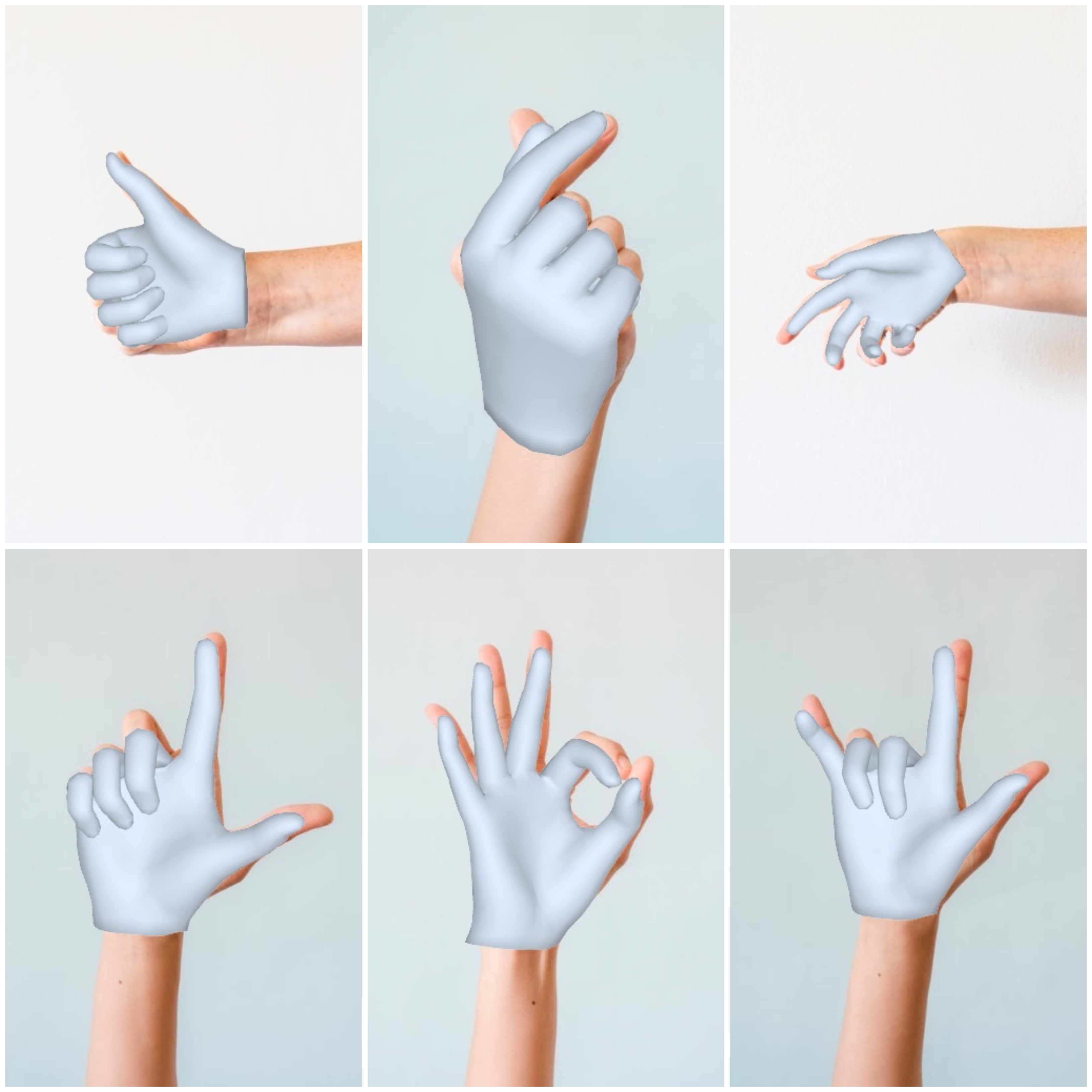}}
  \end{subfigure}
  \begin{subfigure}[t]{0.11\textwidth}
    \fcolorbox{black}{white}{\includegraphics[width=\linewidth]{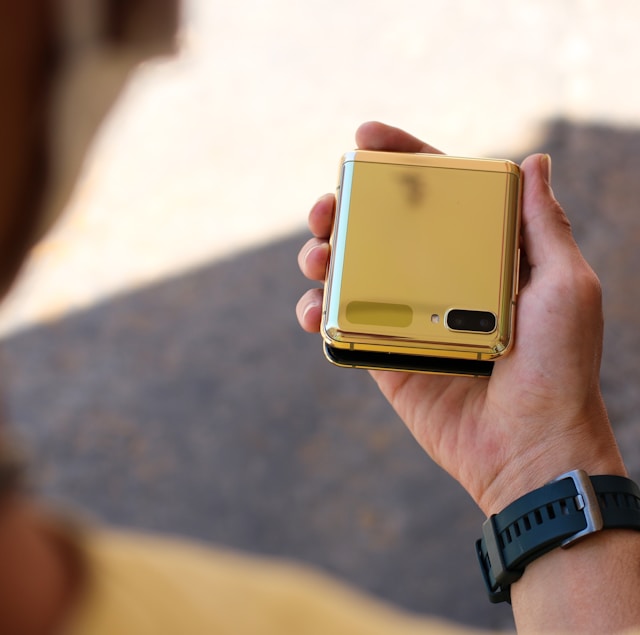}}
  \end{subfigure}
  \begin{subfigure}[t]{0.11\textwidth}
    \fcolorbox{black}{white}{\includegraphics[width=\linewidth]{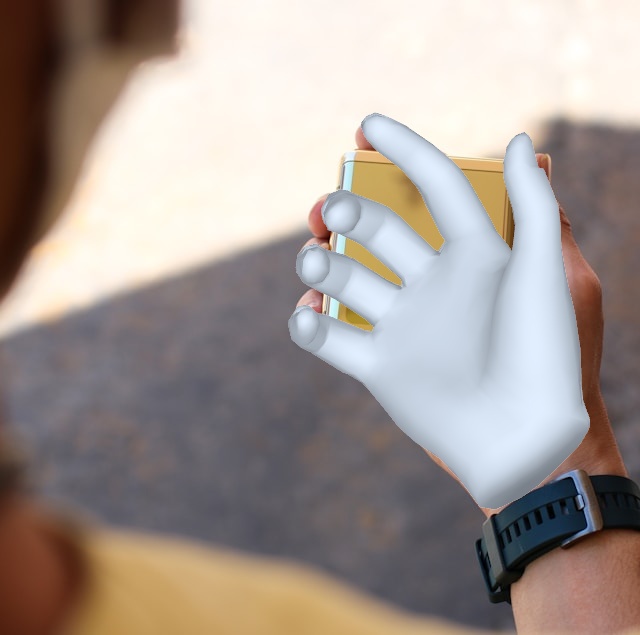}}
  \end{subfigure}
  \begin{subfigure}[t]{0.11\textwidth}
    \fcolorbox{black}{white}{\includegraphics[width=\linewidth]{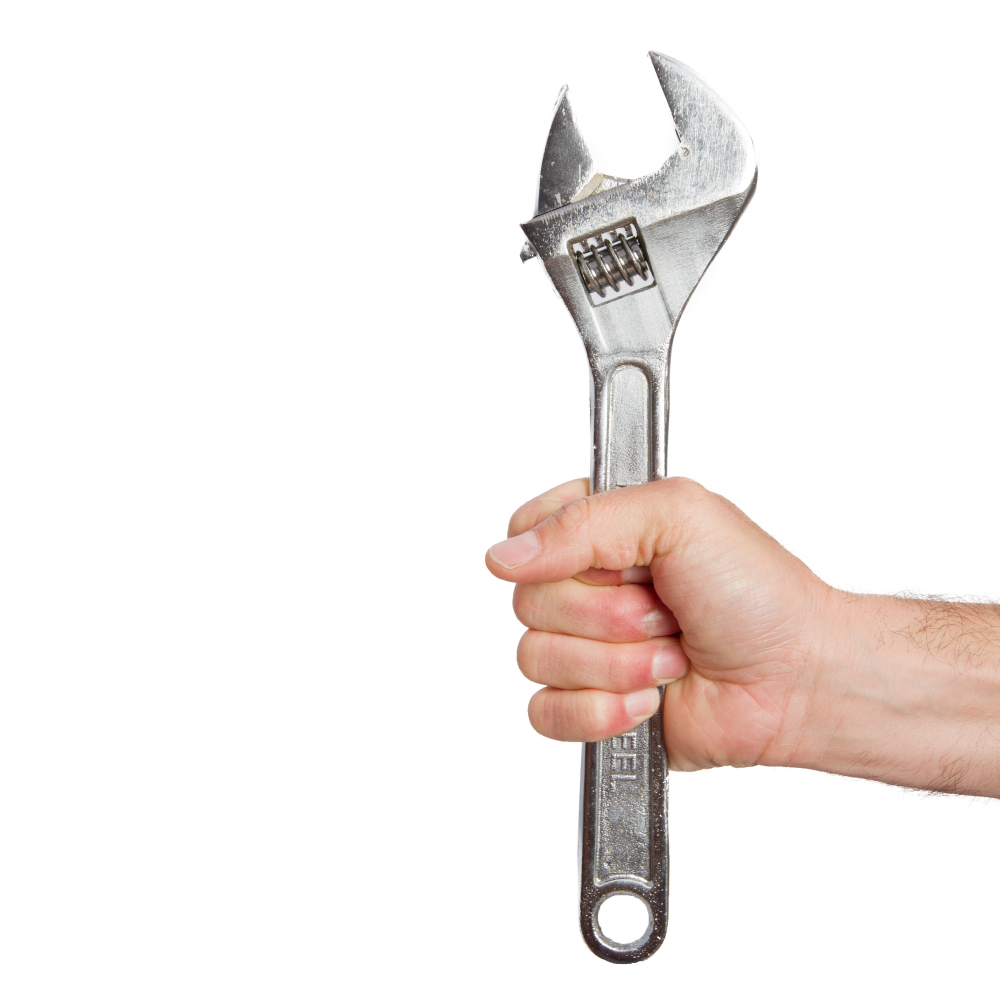}}
  \end{subfigure}
  \begin{subfigure}[t]{0.11\textwidth}
    \fcolorbox{black}{white}{\includegraphics[width=\linewidth]{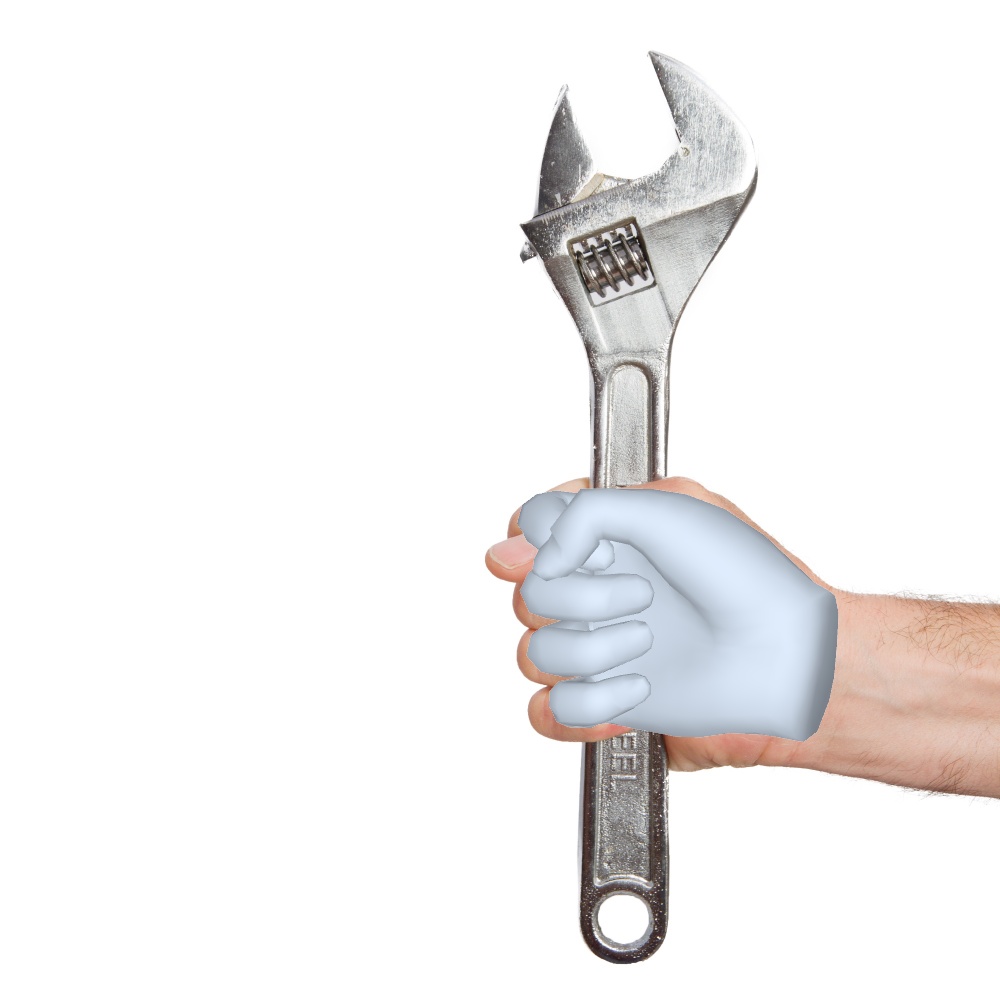}}
  \end{subfigure}
  \caption{\textbf{Real-time monocular 3D hand mesh reconstruction using lightweight student networks.} We present a Knowledge Distillation approach for training lightweight networks that accelerate 3D hand reconstruction without compromising quality. Our best-performing network achieves real-time inference with only a modest drop in reconstruction quality. The Figure showcases qualitative results across challenging scenarios, including mutual occlusion, complex hand poses, and interactions with various objects. }
\end{figure}

A key challenge in computer vision and graphics is estimating the 3D pose and shape of the human hand from visual inputs like images, videos, or depth maps.
Accurate and real-time 3D hand reconstruction is essential for many computer vision and graphics applications such as augmented and virtual reality, human-computer interaction, sign language recognition, and human behavior analysis. 
These applications require precise hand modeling to interpret hand articulations and understand gestures. 
This problem is challenging because human hands are highly articulated, self-occluding, and vary significantly across individuals. 
In recent years, there have been significant advances in hand reconstruction \cite{chatzis2020comprehensive}, which have been made possible with the use of modern Deep Learning techniques \cite{Pavlakos_2024_CVPR}, \cite{li2024hhmr}, \cite{ge20193dhandshapepose}, \cite{Lin_2021_ICCV}, \cite{HandVoxNet2020}, \cite{Aboukhadra_2023_WACV}, \cite{dong2024hamba}, \cite{potamias2025wilor}. 
However, many state-of-the-art hand mesh recovery methods remain computationally expensive~\cite{Pavlakos_2024_CVPR,potamias2025wilor}, as they rely on large-scale networks with transformer-based architectures, making real-time performance difficult to achieve.
\\

To accelerate state-of-the-art 3D hand mesh reconstruction, we propose a Knowledge Distillation (KD) framework for hand mesh reconstruction that transfers knowledge from a large, accurate teacher model to a lightweight student network. 
While applicable to various models, we focus on HaMeR, a transformer-based architecture with a ViT-H backbone. 
Our approach replaces this backbone with more efficient alternatives—such as MobileNet, MobileViT, ConvNeXt, and ResNet—while preserving reconstruction quality. 
During training, the student learns to mimic the teacher at multiple levels: final outputs, intermediate features, or both. 
This distillation strategy enables the student model to achieve near-teacher performance at a fraction of the computational cost, making it suitable for real-time applications on resource-limited devices.

Through practical experiments, we demonstrate that applying Knowledge Distillation, along with using lightweight backbones, leads to improvements in runtime performance with comparable reconstruction accuracy.
This paper makes the following key contributions to the domain of 3D hand reconstruction:

\begin{itemize}
    \item Demonstrating the effectiveness of common distillation losses (output-level and feature-level) for the task of hand mesh reconstruction.
    \item Proposing a lightweight alternative to the state-of-the-art HaMeR model, achieving 35\% of its size, 1.5× faster runtime, and 
    with an accuracy difference of just $0.4mm$.
    \item Analyzing how student architectures respond to different distillation losses, showing that feature-level supervision benefits higher-capacity models, while output-level losses are more effective for smaller or structurally aligned ones.

\end{itemize}
\section{RELATED WORKS}

In recent years, several works have been proposed for monocular 3D hand reconstruction that utilize depth scans \cite{Wu_2018_ECCV,8578976,aboukhadra2024shapegraformer}, RGB frames \cite{rong2021frankmocapmonocular3dwholebody,Lin_2021_ICCV,park2022handoccnet,Pavlakos_2024_CVPR}, or a hybrid RGB-D approach \cite{BMVC2015_77,Sridhar2016RealTimeJT,HandVoxNet2020}. 
These works can be broadly divided into two categories: parametric model-driven approaches and direct mesh recovery approaches. 
Arguably, the primary distinction between these approaches lies in the use of parametric hand models \cite{Romero_2017,10.1145/2816795.2818013,10.1145/3596711.3596797,li2022nimble}. 
These hand models provide prior anatomical constraints to facilitate the reconstruction of the hand.
Parametric model-driven approaches learn a low-dimensional mapping from image features to hand model parameters, which allow them to generate a 3D mesh, while the direct mesh recovery approaches choose to regress the mesh directly without the use of parametric models.

Previous works, such as HAMR \cite{zhang2019endtoendhandmeshrecovery}, introduced a weakly-supervised approach to recover hand mesh and pose from a monocular hand image using silhouette consistency loss.
FrankMocap \cite{rong2021frankmocapmonocular3dwholebody} followed with a modular whole-body pose estimation system, incorporating separate modules for face, hand, and body. 
To address occlusion, HandOccNet~\cite{park2022handoccnet} leveraged transformers, utilizing an attention mechanism to handle occlusions via the correlations between visible and occluded regions.
Supervised learning methods, due to data constraints (i.e., lack of diverse, real-world data), can often fail to generalize well to unconstrained, real-world images.
To address this, ~\cite{hasson2021towards} introduced a learning-free optimization-based approach to reconstructing 3D hand-object interactions, using only RGB video frames. 

Direct mesh recovery methods tend to bypass parametric models by directly regressing 3D hand mesh vertices. 
METRO \cite{Lin2020EndtoEndHP} is a fully transformer-based human pose estimation and mesh reconstruction method, which features a multi-layer transformer encoder.
The core mechanism is Self-attention \cite{10.5555/3295222.3295349}, which enables joint modeling of vertex-vertex and vertex-joint interactions, and it can learn both short and long-range interactions. 
Mesh Graphormer \cite{Lin_2021_ICCV} combines the strengths of self-attention \cite{10.5555/3295222.3295349}, and graph convolutions for human body mesh reconstruction. 
THOR-Net~\cite{Aboukhadra_2023_WACV} introduced the first framework for estimating 3D pose and shape of two hands interacting with an object, along with the texture of the vertices in the resulting hand meshes, through combining the strengths of Graph Convolutional Networks (GCNs) and Transformers with self-supervision. 
Recently, Hamba \cite{dong2024hamba} introduced a graph-guided bi-scanning (GBS) Mamba-based framework \cite{Gu2023MambaLS} for single-view 3D hand mesh reconstruction. 

HaMeR \cite{Pavlakos_2024_CVPR}, the framework we built upon, represents a new state-of-the-art approach for hand mesh reconstruction from a single RGB image. 
It combines a ViT-H~\cite{dosovitskiy2021imageworth16x16words} backbone with a transformer decoder to regress MANO~\cite{Romero_2017} parameters. 
The success of the method can be attributed to scaling the architecture and training data, leading to improved generalization.
Despite outperforming all previous RGB-based approaches, scaling the HaMeR network came at the cost of runtime and resource efficiency, which makes it unsuitable for real-time applications.
Therefore, in our work, we aim to improve runtime and resource efficiency by utilizing HaMeR as a reference for training lighter networks.



\subsection{Knowledge Distillation for Acceleration}

Knowledge Distillation (KD) \cite{hinton2015distilling} is a famous technique for reducing network complexity that transfers the knowledge from large teacher networks to smaller student networks.
Distillation was used recently to compress and accelerate Large Language Models (LLMs)~\cite{ko2024distillm,gu2024minillm,hsieh2023distilling}. 
On the other hand, KD has been only marginally explored in the context of 3D reconstruction and pose estimation problems. 
Works such as \cite{zhang2019fast} and \cite{hwang2020lightweight} focus on full-body pose tasks, distilling 2D or 3D joint information without addressing dense geometric structures. 
~\cite{zhang2020} and, more recently ~\cite{capistrano2025}, apply KD to hand pose estimation, transferring pose-level knowledge into compact models. 
In contrast, our method applies both feature and output distillation to compress a full mesh reconstruction model, preserving high-quality geometry beyond joint estimation.




\section{METHOD}


Our goal is to develop a fast, lightweight, and efficient method for parametric 3D hand reconstruction that matches the performance of slower, resource-intensive models. 
While our framework is designed to generalize across different reconstruction algorithms, we focus on accelerating the state-of-the-art HaMeR \cite{Pavlakos_2024_CVPR} model without loss of generality.
We achieve this task by training smaller and lighter student networks that learn from the original high-capacity teacher network (i.e., HaMeR) through the use of Knowledge Distillation. 
At a high level, the framework consists of the original HaMeR as a teacher network \textbf{\textit{T}} with frozen weights, and a smaller and lighter network that acts as a student \textbf{\textit{S}}. 

\begin{figure*}[t]
    \centering
    \includegraphics[width=0.95\linewidth]{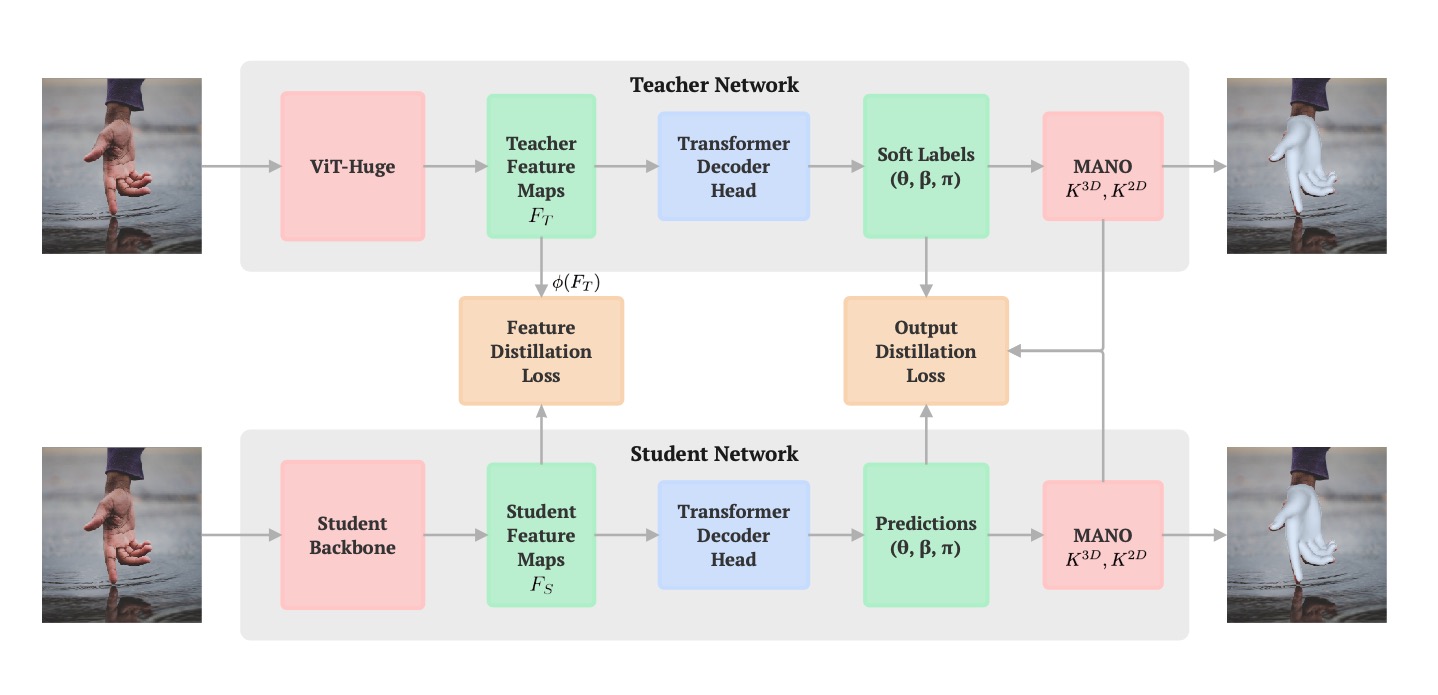}
    \caption{High-level overview of the teacher-student architecture, only relevant distillation losses are used depending on the KD level. \( \phi(F_T) \) refers to a 1x1 convolution to project the dimensions of the teacher's features to match those of the student's for feature-level distillation.
    The teacher network is used only during training; at inference time, only the trained student network is used.
    }
    \label{fig:combined-level-distillation}
\end{figure*}

\subsection{Teacher Model (HaMeR)} \label{sec:architecture-hamer}

HaMeR \cite{Pavlakos_2024_CVPR} features a fully transformer-based design, in particular, it adopts a Huge Vision Transformer (ViT-H) \cite{dosovitskiy2021imageworth16x16words} as its backbone.
The output token sequence (i.e. image features) is fed into a transformer-decoder head responsible for regressing the MANO parameters $(\theta, \beta)$ and camera parameters $\pi$, where $\theta \in \mathbb{R}^{48}, \beta \in \mathbb{R}^{10}$. The transformer head utilizes self-attention and processes a single token while cross-attending to the backbone output tokens \cite{Pavlakos_2024_CVPR}.
The regressed parameters are fed into the MANO head to obtain a 3D mesh of the hand along with the 3D joint locations $K_T^{3D}$ for 21 keypoints, where $K_T^{3D} \in \mathbb{R}^{3 \times N}$, and $N = 21$. 
The camera parameters includes a 3D translation vector $(t \in \mathbb{R}^3)$ necessary to project the 3D mesh and the 3D joint locations onto the image plane and in obtaining the 2D joint locations $K_T^{2D} \in \mathbb{R}^{2 \times N}$.

In order to utilize datasets with no 3D annotations, HaMeR applies a re-projection loss \(L_{2D}\) between the 2D projection of the 3D predicted joint locations and 2D ground-truth joint locations. 
The Keypoint 3D Loss \(L_{3D}\) measures the distance between the predicted 3D joint locations and the ground-truth joint locations, in order to encourage the model to reconstruct the correct 3D hand pose.
For datasets that provide MANO~\cite{Romero_2017} annotations, the MANO parameter loss \(L_{MANO}\) penalizes the difference between the predicted MANO parameters and the ground-truth parameters, ensuring the model learns to predict the correct parameters. 
The combined loss of \(L_{2D}\), \(L_{3D}\), and \(L_{MANO}\) will be referred to as $\mathcal{L}_{GT}$.


\subsection{Student Model}

The student network retains the overall architecture of the teacher model; however, in our work, we replace the larger ViT-Huge backbone \cite{dosovitskiy2021imageworth16x16words} with smaller networks, namely, \textit{MobileNet} \cite{Howard2017MobileNetsEC}, \textit{MobileViT} \cite{Mehta2021MobileViTLG}, \textit{ConvNeXt} \cite{Liu2022ACF}, and \textit{ResNet} \cite{He2015DeepRL}.
We utilize these networks as backbones in our student network $S$, aiming to evaluate the trade-offs between reconstruction accuracy and inference speed. 
These choices represent a few design philosophies, such as lower parameter counts, depthwise convolutions \cite{Howard2017MobileNetsEC}, lightweight hybrid transformer-based models \cite{Mehta2021MobileViTLG}, and residual connections \cite{He2015DeepRL}. 
Furthermore, the criteria for choosing this subset of backbones is their relative performance on the ImageNet-1K dataset \cite{deng2009imagenet}. 
Additionally, we considered their number of parameters stated in the torchvision library, and their runtime efficiency. 

\subsection{Knowledge Distillation}
\label{sec:kd}

During the training process, an input image is passed through both the student \textbf{\textit{S}} and the teacher network \textit{\textbf{T}}. Both \textit{\textbf{T}} and \textit{\textbf{S}} produce intermediate feature maps (i.e, ${F_T}$ and {${F_S}$) and final predictions (i.e, $\hat{Y}_T$ and $\hat{Y}_S$) where $\hat{Y} = \{K^{3D},K^{2D},\theta, \beta, \pi\}$.
We enforce similarity between the predictions on three different levels: output-level, feature-level, and combined distillation. We then evaluate each approach to determine which performs best. 

In the first approach, we calculate distillation losses between $\hat{Y}_T$ and $\hat{Y}_S$ (i.e., \textit{output-level distillation}). 
In the second approach we calculate distillation losses between ${F_T}$ and ${F_S}$ (i.e, \textit{feature-level distillation}), and finally, between both ${F_T}$ and ${F_S}$, and $\hat{Y}_T$ and $\hat{Y}_S$ (i.e., \textit{combined distillation}). 
Figure \ref{fig:combined-level-distillation} shows the different distillation methods.
By enforcing these similarities, we allow the smaller student model to learn how the bigger teacher model learns and represents the inputs internally, at the same time decreasing output latency due to the smaller size of \textit{\textbf{S}}. We finally evaluate each distillation scheme to select the best approach.
\paragraph{Output-level distillation}
In the case of \textit{output-level distillation}, to enforce similarity between $\hat{Y}_T$ and $\hat{Y}_S$, we use the same loss terms from the original network (specified in Section \ref{sec:architecture-hamer}) but substitute the ground truth values for the teacher's predictions. 
The total loss for output-level distillation includes the original losses $\mathcal{L_{GT}}$ on the ground truths in addition to the losses between the student's predictions $\hat{Y}_S$ and the teacher's predictions $\hat{Y}_T$. 
We also use $\lambda_{KD}$ to control how much distillation losses contribute to the total loss.

\begin{equation}
\mathcal{L}_{total,out} = \mathcal{L}_{GT} + \lambda_{KD}\mathcal{L}_{KD, out}  
\end{equation}

\begin{equation}
\resizebox{\columnwidth}{!}{$
\mathcal{L}_{KD,out}
= \lVert \hat{K}^{3D}_S - \hat{K}^{3D}_T \rVert_2^2
+ \lVert \hat{K}^{2D}_S - \hat{K}^{2D}_T \rVert_2^2
+ \lVert \hat{\Theta}_S - \hat{\Theta}_T \rVert_2^2
$}
\end{equation}



\paragraph{Feature-level distillation}
Knowledge Distillation on the feature maps (i.e., \textit{feature-level distillation}) enables the student \textbf{\textit{S}} to learn feature representations similar to those of the teacher network \textbf{\textit{T}}. Since the output feature maps $F_S$ and $F_T$ may differ in channel dimensionality, we apply a learnable 1x1 convolution $\phi$ to project the feature dimensions of $F_T$ to match those of $F_S$. 
Additionally, we align the spatial dimensions using bilinear interpolation when they differ.
The total loss for the feature-level distillation includes losses on the ground-truth data $\mathcal{L_{GT}}$, and between teacher and student features (i.e., ${F_T}$ and ${F_S}$ respectively): 
$\mathcal{L}_{KD, feat} = ||F_S- \phi(F_T)||^2_2$.
$\gamma_{feat}$ is used to scale the feature-distillation loss, while $\lambda_{KD}$ controls the contribution of the distillation loss to the total loss.

\begin{equation}   
\mathcal{L}_{total,feat} = \mathcal{L}_{GT} + \lambda_{KD}(\gamma_{feat}\times\mathcal{L}_{KD, feat})
\end{equation}

\paragraph{Combined distillation}
Finally, the motivation behind \textit{combined distillation} is to guide not just the final representations but also the intermediate representations simultaneously, in hopes of achieving better results by combining the merits of output distillation and feature distillation.

\begin{equation}
\mathcal{L}_{total,comb} = \mathcal{L}_{GT} + \lambda_{KD}(\mathcal{L}_{KD, out} + \gamma_{feat}\times\mathcal{L}_{KD, feat})
\end{equation}

\subsection{Datasets}

The reason behind the state-of-the-art performance of HaMeR lies in its scaling of training data and the learning capacity of the network architecture. The dataset used for training HaMeR consists of approximately 2.7 million annotated samples, combined from different sources.
FreiHAND \cite{Zimmermann_2019_ICCV}, HO3D \cite{Hampali_2020_CVPR}, H2O3D \cite{Hampali_2020_CVPR}, InterHand2.6M \cite{moon2020interhand26mdatasetbaseline3d}, MTC \cite{Xiang2018MonocularTC}, DexYCB \cite{Chao_2021_CVPR} provide 3D annotations along with RHD \cite{Zimmermann_2017_ICCV} which is a synthetic dataset, while COCO WholeBody \cite{jin2020whole}, Haple-FullBody \cite{fang2022alphaposewholebodyregionalmultiperson}, MPII + NZSL \cite{simon2017hand} contribute with 2D annotations.
The unification of various hand pose datasets into a consistent format ready to be used is a crucial contribution from the HaMeR team, allowing large-scale supervised training with minimal pre-processing.
Accordingly, we adopt the same combined dataset and sampling distributions used in HaMeR for training all of our networks.
We evaluate and provide quantitative results on the HO3D-v2 \cite{Hampali_2020_CVPR} dataset, allowing us to compare our results directly with HaMeR and other baselines. 
The dataset serves as an ideal evaluation set as it offers challenging and realistic hand-object interaction scenarios.
\section{EXPERIMENTS AND RESULTS}

In this section, we present a quantitative evaluation of our proposed KD approach on the HO3D-v2 dataset \cite{Hampali_2020_CVPR}. 
We use standard evaluation metrics, including \textit{PA-MPJPE} ($\mathbf{J}_{err}$), and \textit{PA-MPVPE} ($\mathbf{V}_{err}$) both being in \textit{mm}, and \textit{F@5mm}, and \textit{F@15mm}. 
The official HO3D-v2 competition website provides more details on the metrics used for evaluation\footnote{\url{https://codalab.lisn.upsaclay.fr/competitions/4318#learn_the_details}}.
We also comment on the performance of the model with respect to its number of parameters and FPS, as shown in Table~\ref{tab:parameters-to-runtime} and Figure~\ref{fig:accuracy_vs_fps}. 
All experiments and FPS calculations are evaluated on a consumer-grade \textit{RTX-4060 Ti} GPU.


\subsection{Ablation Study}

We start by establishing performance baselines using lightweight backbones without any distillation.
Subsequently, we compare the different distillation methods to assess their impact on performance. 


\subsubsection{Baseline Performance without Distillation}
\label{sec:baselines}
Prior to applying any distillation, we establish baselines by replacing HaMeR's original ViT-H backbone~\cite{dosovitskiy2021imageworth16x16words,Pavlakos_2024_CVPR} with a range of lightweight models~\cite{Howard2017MobileNetsEC,Mehta2021MobileViTLG,He2015DeepRL,Liu2022ACF}. 
Each model is initialized with ImageNet-pretrained weights and trained only using ground-truth supervision, without the teacher model's guidance.

The baseline experiments, shown in Table~\ref{tab:base_results}, highlight a clear trade-off between model complexity and performance. 
The original HaMeR (671M parameters) produces the best results across all metrics, but runs at only 27 FPS. 
In contrast, lighter models like MobileViT-S~\cite{Mehta2021MobileViTLG} reduce parameter count by \textit{6x} and improve the inference speed by \textit{1.55x}, but with a \textit{1.6mm} drop in accuracy. 
ResNet-50~\cite{He2015DeepRL} offers promising results with only 10\% of the parameter count, and \textit{1.85x} speed-up, at just a \textit{1mm} difference in accuracy. 
ConvNeXt-L~\cite{Liu2022ACF}, arguably, offers the best trade-off between compression and accuracy, with a \textit{1.48x} increase in FPS and about \textit{64\%} reduction in complexity, performs only \textit{0.6mm} worse than HaMeR. 
These results demonstrate promising trade-offs and indicate that data scaling may be more beneficial than increasing model complexity alone.


\begin{figure}[t]
\centering
\begin{tikzpicture}
\begin{axis}[
    width=\columnwidth,  
    height=0.53\textwidth,  
    xlabel={PA-MPJPE (mm) $\downarrow$},
    ylabel={FPS $\uparrow$},
    x dir=reverse,
    enlargelimits=0.15,
    grid=both,
    legend style={
    at={(0.5,-0.13)},  
    anchor=north,
    legend columns=3,
    column sep=0.2cm,
    align=left,
    draw=none,
    font=\scriptsize,
    },
    scatter/classes={
        MobileNet={mark=*,draw=black,fill=gray},
        MobileViT={mark=*,draw=black,fill=red},
        ResNet50={mark=*,draw=black,fill=pink},
        ResNet101={mark=*,draw=black,fill=blue},
        ConvNeXt={mark=*,draw=black,fill=green},
        HaMeR={mark=*,draw=black,fill=yellow}
    }
]

\addplot[
    scatter,
    only marks,
    scatter src=explicit symbolic,
    point meta=explicit symbolic,
    every node near coord/.append style={
    font=\small,
    yshift=10pt,          
    fill=none,
    fill opacity=0.8,
    text opacity=1
    },
    visualization depends on=\thisrow{size} \as \perpointmarksize,
    scatter/@pre marker code/.append style={
        /tikz/mark size=\perpointmarksize,
    }
]
table[meta=type] {FastHaMeR/figs/accuracy-fps-data.dat};

\legend{
MobileNet-Large,
MobileViT-Small,
ResNet50,
ResNet101,
ConvNeXt-Large,
ViT-Huge (HaMeR)
}

\end{axis}
\end{tikzpicture}
\caption{Trade-off between model accuracy (PA-MPJPE), speed (FPS), and parameter size (circle size). 
The Figure shows that the most accurate model is HaMeR; however, it shows that other alternatives give close performance with fewer resources and better runtime. 
ConvNeXt-L with feature distillation falls right behind HaMeR in our experiments with 1.48x FPS boost and 64\% reduction in size.
}
\label{fig:accuracy_vs_fps}
\end{figure}
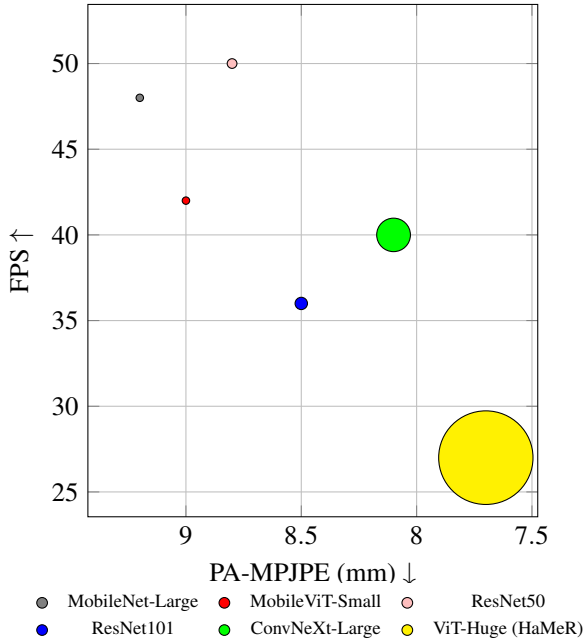

\begin{table}[b]
\normalsize
\centering
\caption{The number of networks' parameters in millions, along with the runtime of different backbone configurations in Frames-per-Second (FPS). 
All configurations were tested on a consumer-grade RTX 4060 Ti GPU. 
We also report the number of floating point operations (GFLOPs) as stated in the torchvision library.}
\resizebox{0.8\columnwidth}{!}{%
\begin{tabular}{@{}lccc@{}}
\toprule
\textbf{Backbone} &
  \textbf{\begin{tabular}[c]{@{}c@{}}Total\\ Params (M)\end{tabular}} &
  \textbf{FPS $\uparrow$} &
  \textbf{GFLOPs} \\ 
\midrule
MobileNet-L & 42.0 & 48 & 0.2 \\ 
MobileViT-S & 42.0 & 42 & 0.5 \\ 
ResNet-50 & 69.3 & 50 & 4.1 \\ 
ResNet-101 & 88.3 & 36 & 7.8 \\ 
ConvNeXt-L & 240  & 40 & 34.4 \\ 
ViT-H & 671  & 27 & 167.3 \\ 
\bottomrule
\end{tabular}%
}
\label{tab:parameters-to-runtime}
\end{table}

\begin{table}[]
\small
\setlength{\tabcolsep}{4pt}
\centering
\caption{Performance evaluation after training different backbones without KD to set baselines.
The experiments are sorted in ascending order of the total parameter size of the network.}
\resizebox{0.9\columnwidth}{!}{%
\begin{tabular}{@{}lccc@{}}
\toprule
\textbf{Backbone} &
    \textbf{\begin{tabular}[c]{@{}c@{}}$\mathbf{J}_{err}$ $\downarrow$\end{tabular}} & 
    \textbf{\begin{tabular}[c]{@{}c@{}}$\mathbf{V}_{err}$ $\downarrow$\end{tabular}} & 
  \textbf{\begin{tabular}[c]{@{}c@{}}F@5.0 / F@15.0 $\uparrow$\end{tabular}} \\ 
\midrule
MobileNet-L      & 9.2 & 9.3 & 0.532 / 0.967 \\
MobileViT-S      & 9.5 & 9.5 & 0.515 / 0.966 \\
ResNet-50        & 8.8 & 8.9 & 0.563 / 0.971 \\
ResNet-101       & 8.8 & 9.0 & 0.558 / 0.971 \\
\rowcolor{lightorange} ConvNext-L & 8.3 & 8.5 & 0.599 / 0.976 \\ 
\rowcolor{lightred} ViT-H (HaMeR) & 7.7 & 7.9 & 0.635 / 0.980 \\
\bottomrule
\end{tabular}%
}
\label{tab:base_results}
\end{table}



\begin{figure*}[t]
    \centering
    \setlength{\tabcolsep}{2pt} 
    \resizebox{0.9\textwidth}{!}{
    \begin{tabular}{ccccccc}
        
                    

        \begin{subfigure}{0.14\textwidth}
            
            \caption*{Input\vspace{1mm}}
            \includegraphics[width=\linewidth,height=\linewidth,keepaspectratio=false]
            {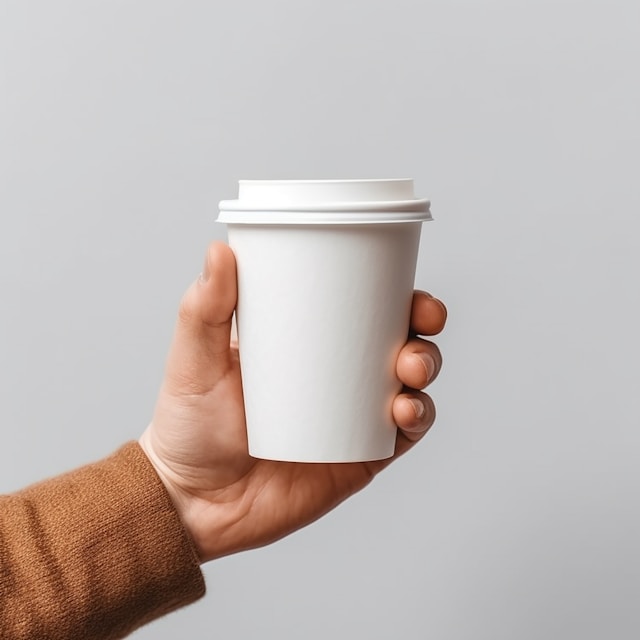}
        \end{subfigure} &
        \begin{subfigure}{0.14\textwidth}
            \caption*{HaMeR\vspace{1mm}}
            
            \includegraphics[width=\linewidth,height=\linewidth,keepaspectratio=false]{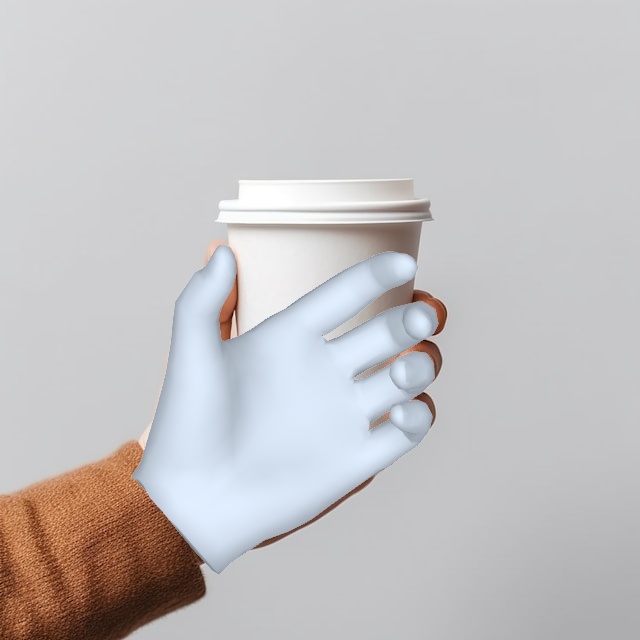}
        \end{subfigure} &
        \begin{subfigure}{0.14\textwidth}
            \caption*{Top View\vspace{1mm}}
            
            \includegraphics[width=\linewidth,height=\linewidth,keepaspectratio=false]{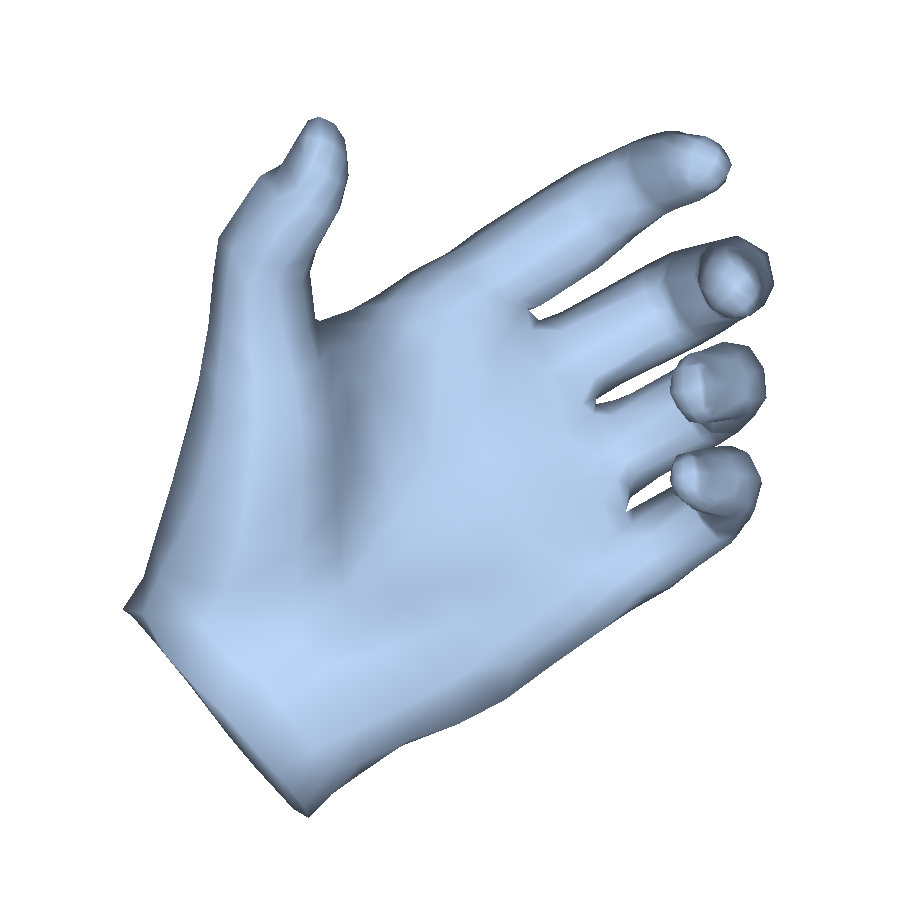}
        \end{subfigure} &
        \begin{subfigure}{0.14\textwidth}
            \caption*{Side View\vspace{1mm}}
            
            \includegraphics[width=\linewidth,height=\linewidth,keepaspectratio=false]{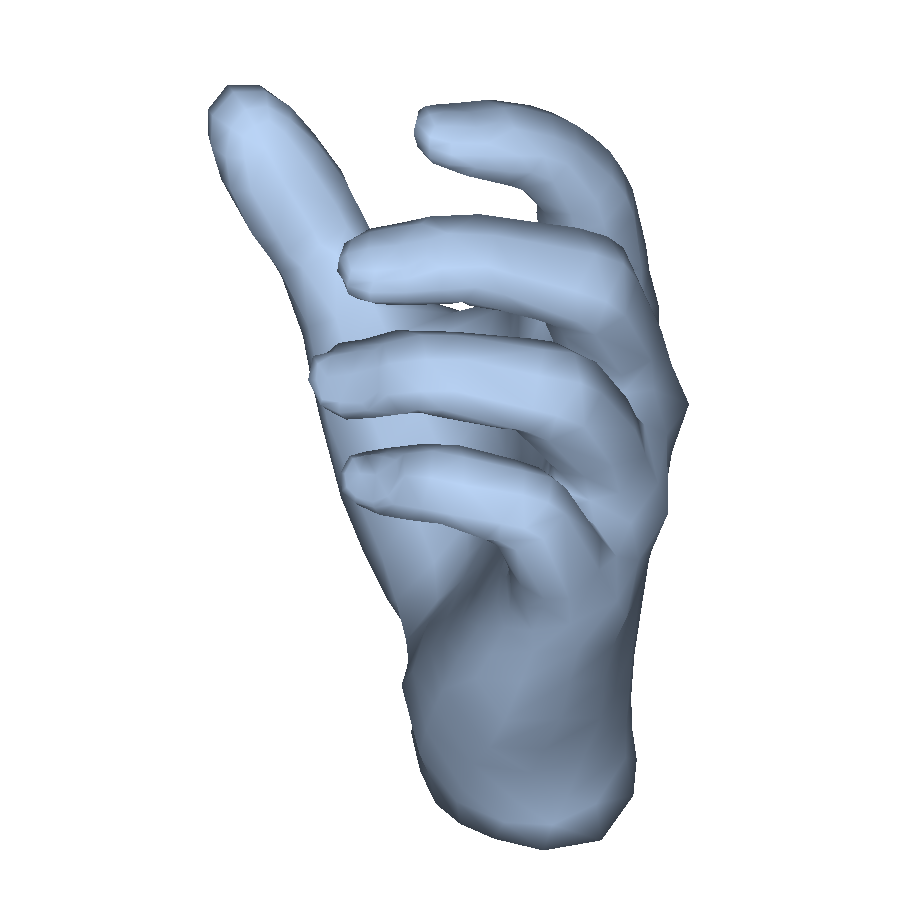}
        \end{subfigure} &
        \begin{subfigure}{0.14\textwidth}
        \caption*{Ours\vspace{1mm}}
            
            \includegraphics[width=\linewidth,height=\linewidth,keepaspectratio=false]{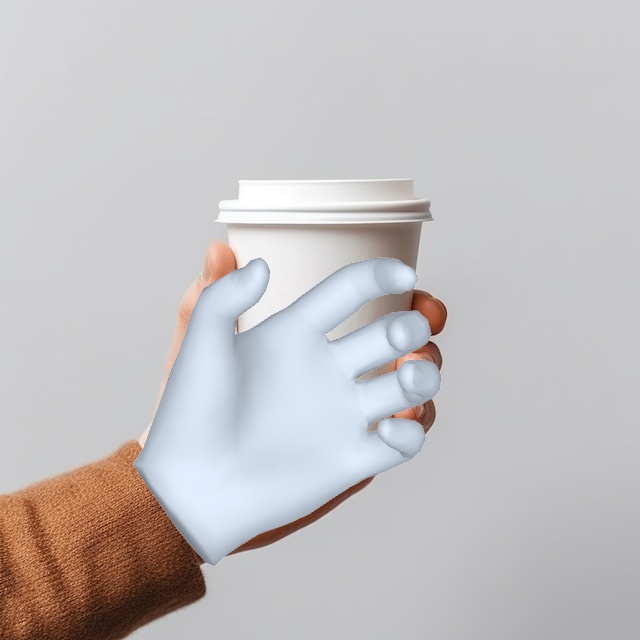}
        \end{subfigure} &
        \begin{subfigure}{0.14\textwidth}
        \caption*{Top View\vspace{1mm}}
            
            \includegraphics[width=\linewidth,height=\linewidth,keepaspectratio=false]{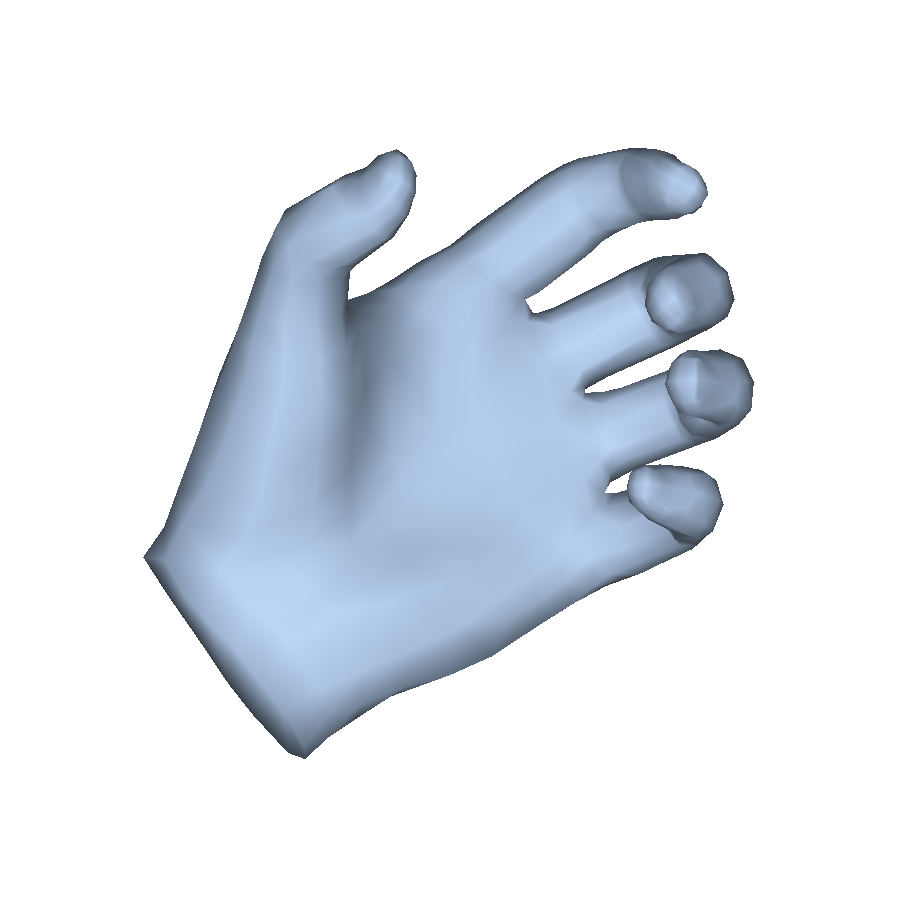}
        \end{subfigure} &
        \begin{subfigure}{0.14\textwidth}
        \caption*{Side View\vspace{1mm}}
            
            \includegraphics[width=\linewidth,height=\linewidth,keepaspectratio=false]{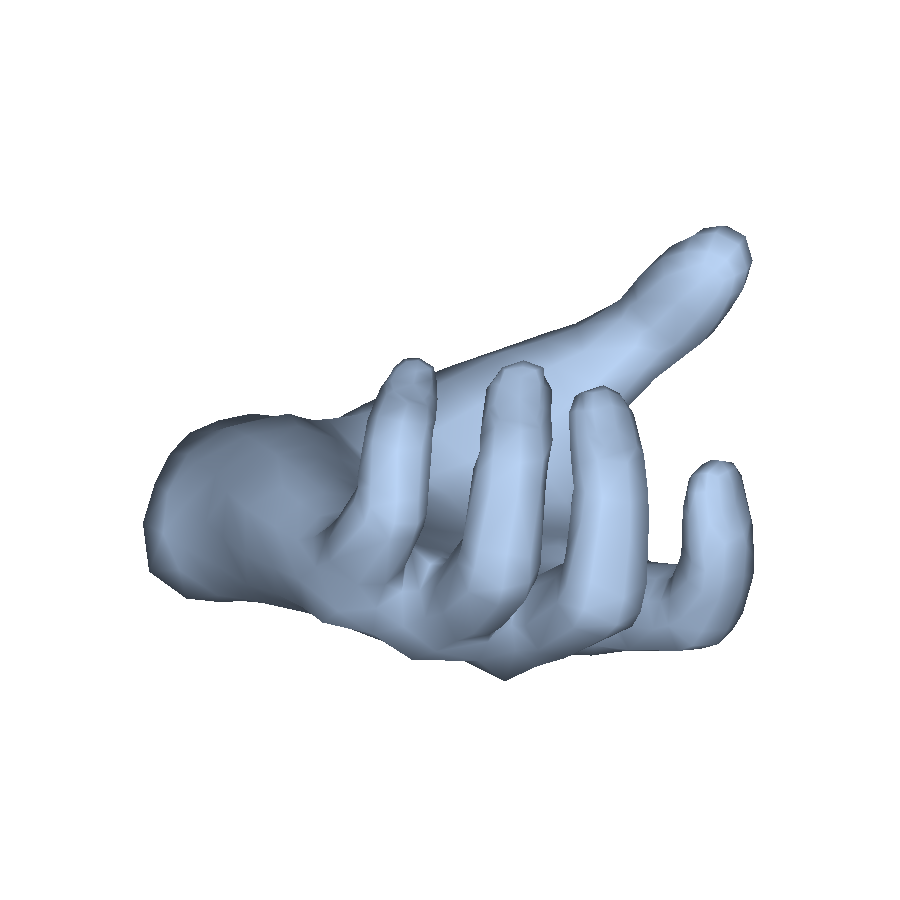}
        \end{subfigure} \\

        \begin{subfigure}{0.14\textwidth}
            
            \includegraphics[width=\linewidth,height=\linewidth,keepaspectratio=false]
            {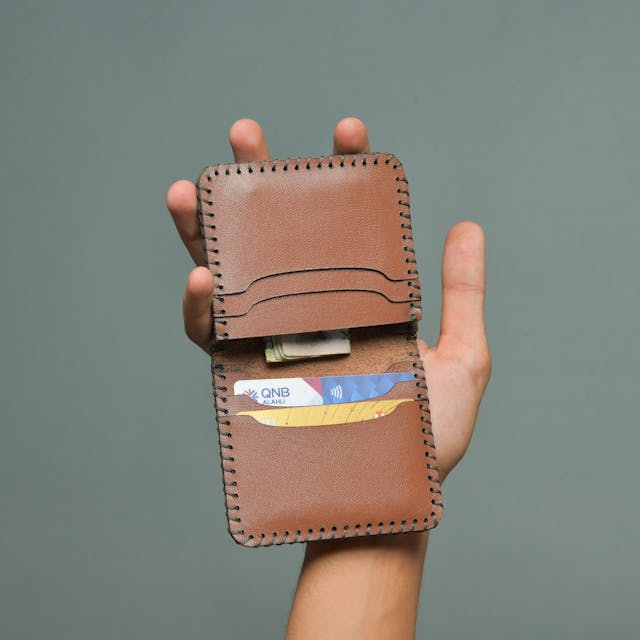}
        \end{subfigure} &
        \begin{subfigure}{0.14\textwidth}
            
            \includegraphics[width=\linewidth,height=\linewidth,keepaspectratio=false]{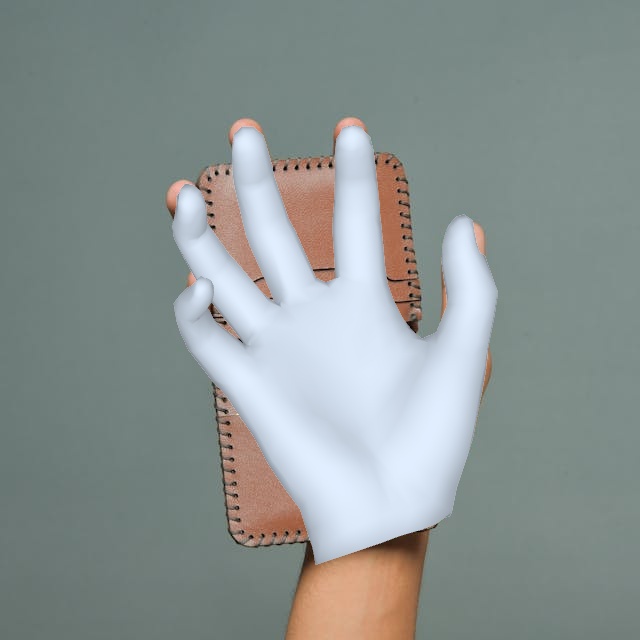}
        \end{subfigure} &
        \begin{subfigure}{0.14\textwidth}
            
            \includegraphics[width=\linewidth,height=\linewidth,keepaspectratio=false]{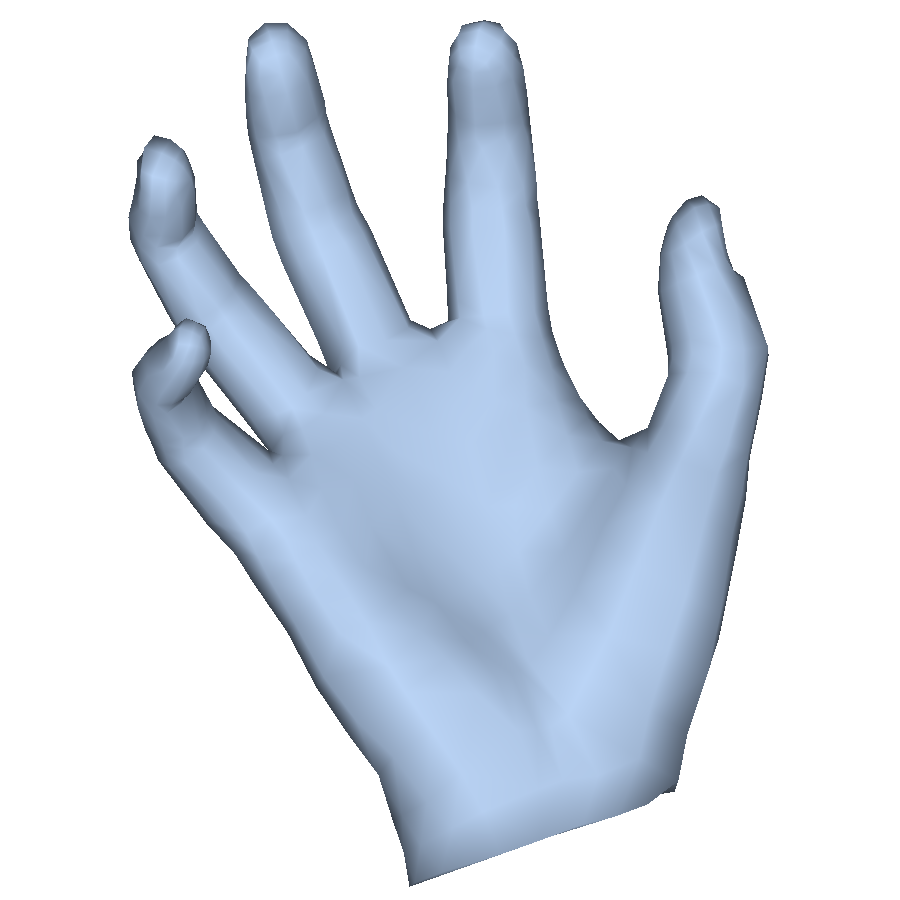}
        \end{subfigure} &
        \begin{subfigure}{0.14\textwidth}
            
            \includegraphics[width=\linewidth,height=\linewidth,keepaspectratio=false]{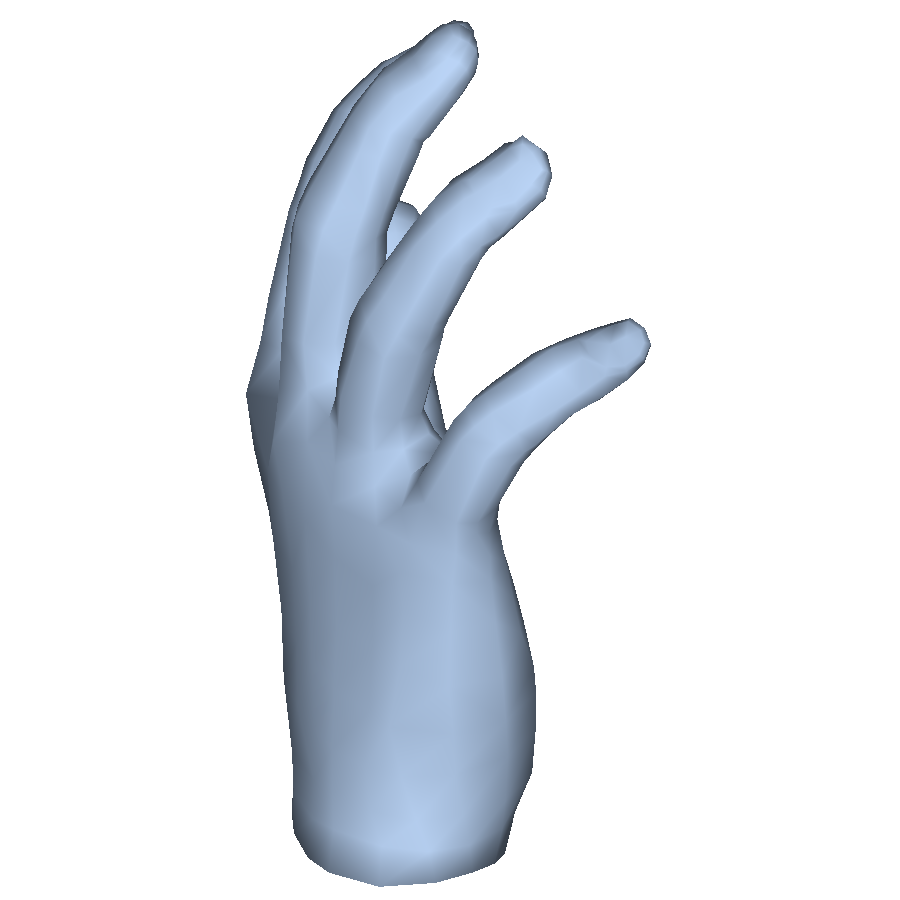}
        \end{subfigure} &
        \begin{subfigure}{0.14\textwidth}
            
            \includegraphics[width=\linewidth,height=\linewidth,keepaspectratio=false]{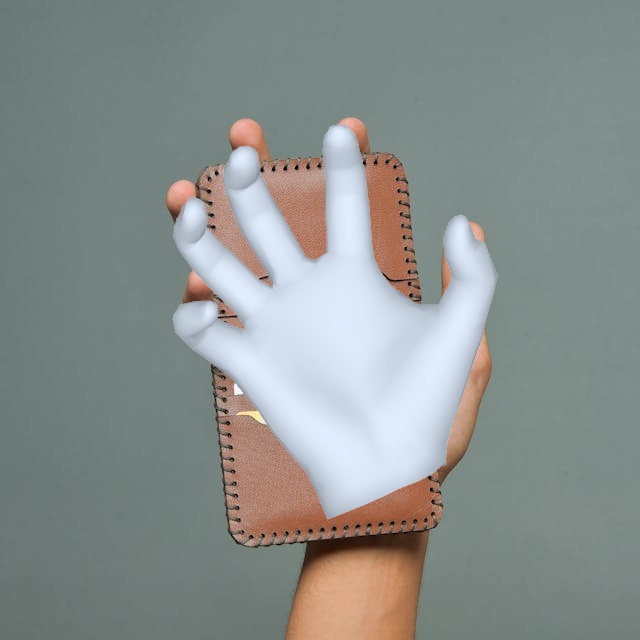}
        \end{subfigure} &
        \begin{subfigure}{0.14\textwidth}
            
            \includegraphics[width=\linewidth,height=\linewidth,keepaspectratio=false]{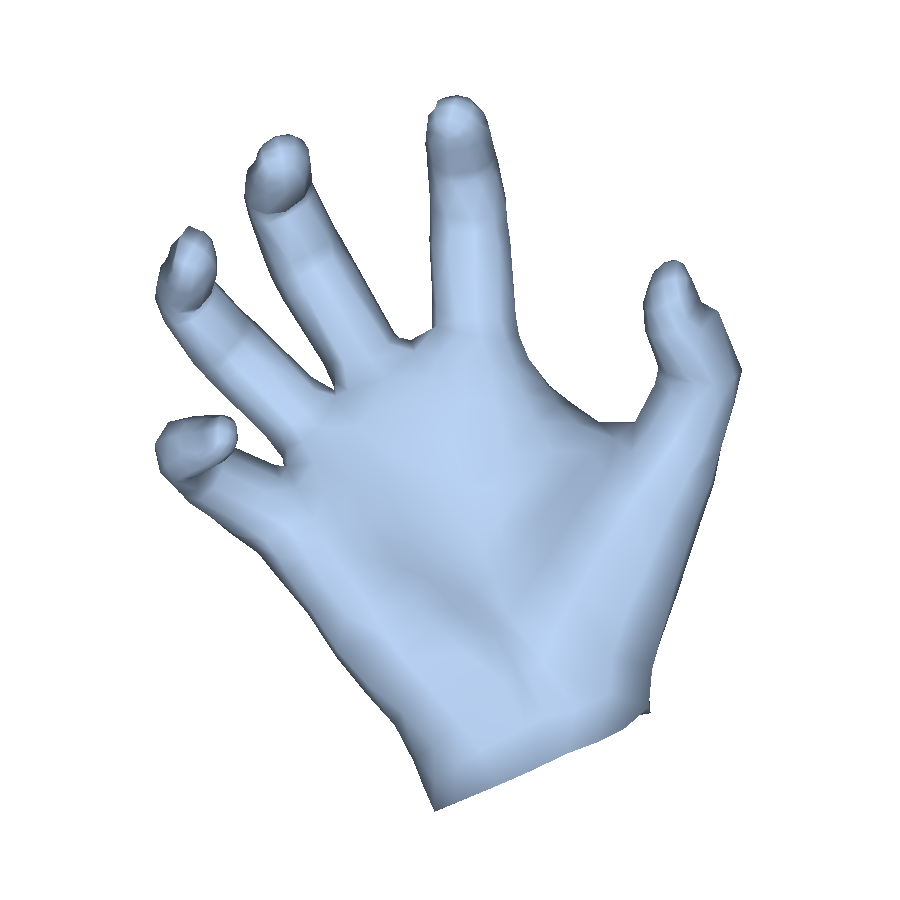}
        \end{subfigure} &
        \begin{subfigure}{0.14\textwidth}
            
            \includegraphics[width=\linewidth,height=\linewidth,keepaspectratio=false]{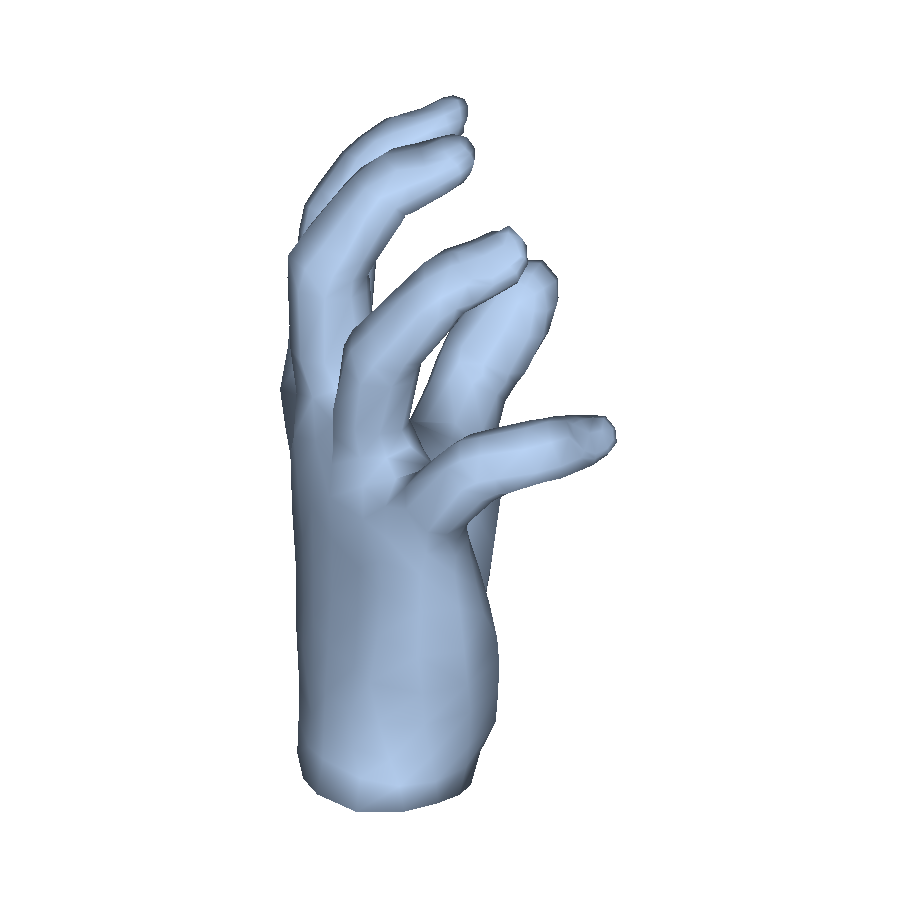}
        \end{subfigure} \\





        \begin{subfigure}{0.14\textwidth}
            \includegraphics[width=\linewidth,height=\linewidth,keepaspectratio=false]{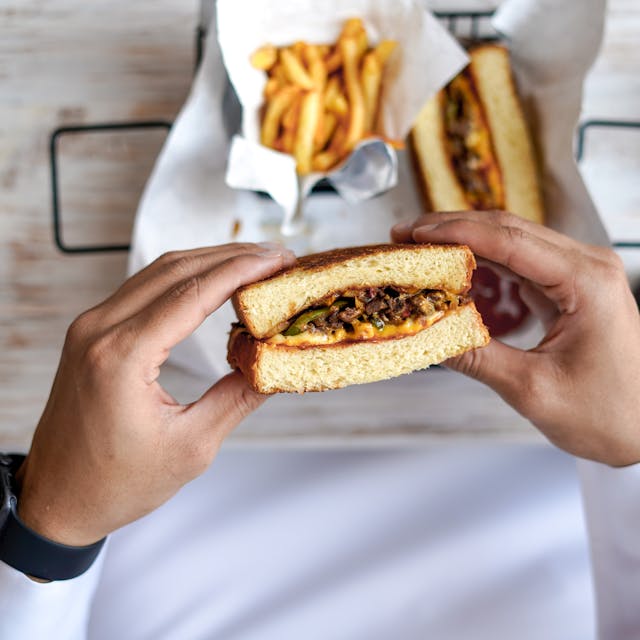}
        \end{subfigure} &
        \begin{subfigure}{0.14\textwidth}
            \includegraphics[width=\linewidth,height=\linewidth,keepaspectratio=false]{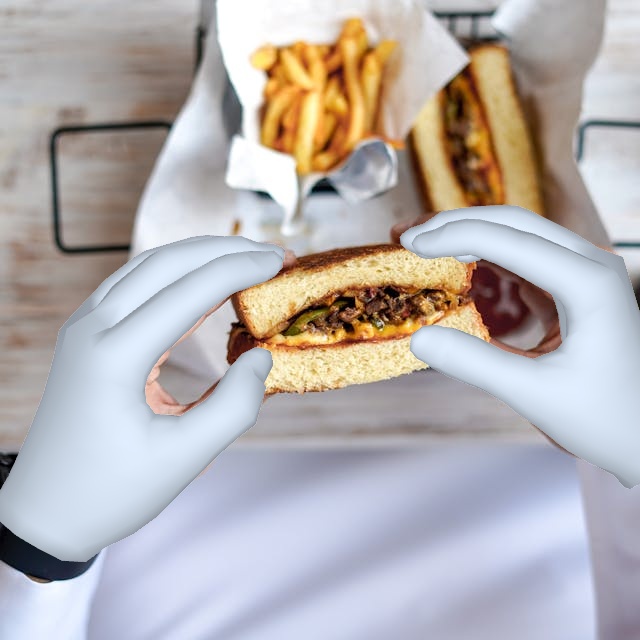}
        \end{subfigure} &
        \begin{subfigure}{0.14\textwidth}
            \includegraphics[width=\linewidth,height=\linewidth,keepaspectratio=false]{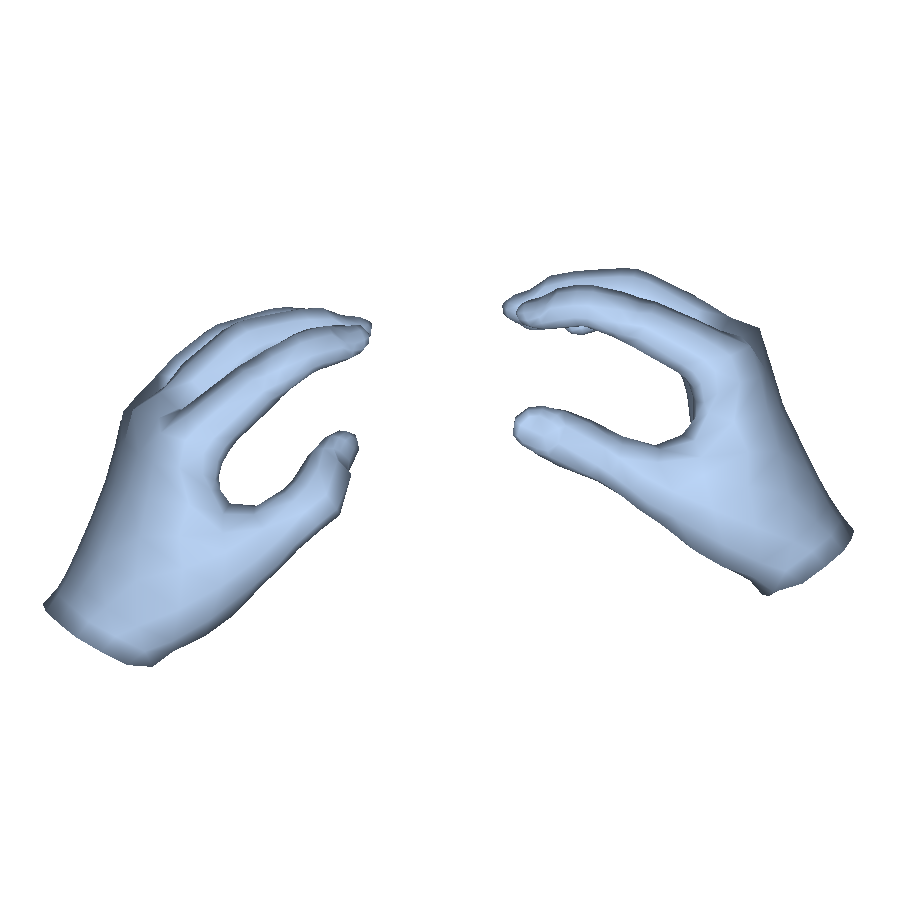}
        \end{subfigure} &
        \begin{subfigure}{0.14\textwidth}
            \includegraphics[width=\linewidth,height=\linewidth,keepaspectratio=false]{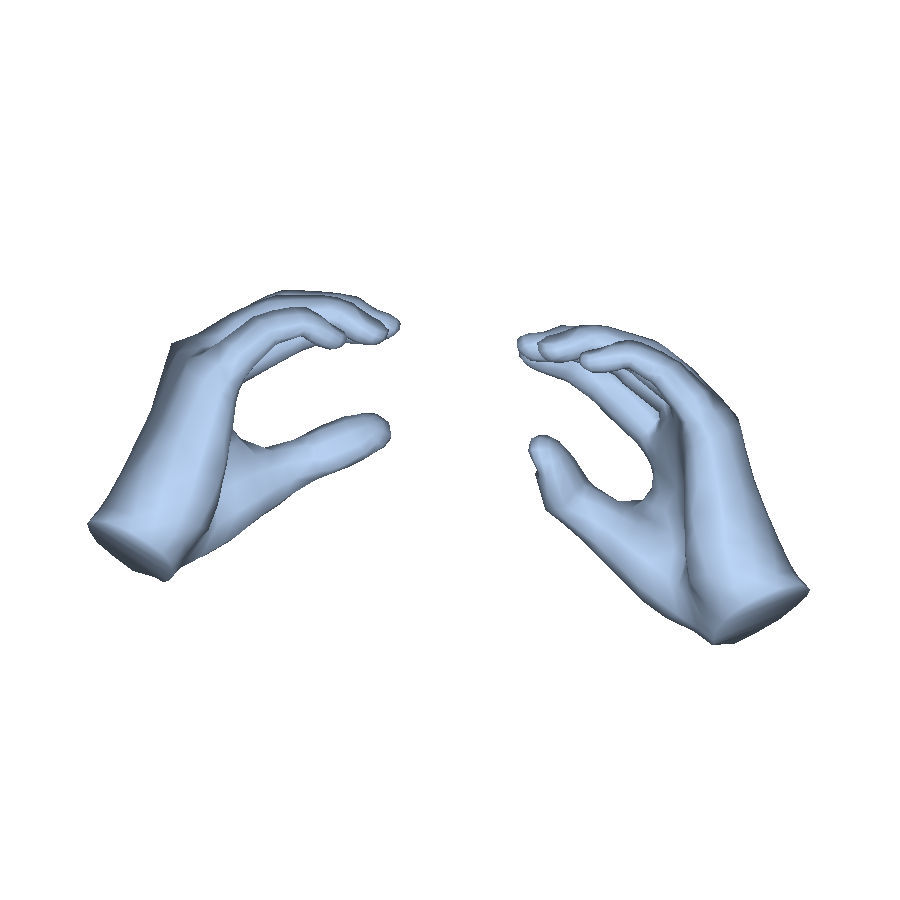}
        \end{subfigure} &
        \begin{subfigure}{0.14\textwidth}
            \includegraphics[width=\linewidth,height=\linewidth,keepaspectratio=false]{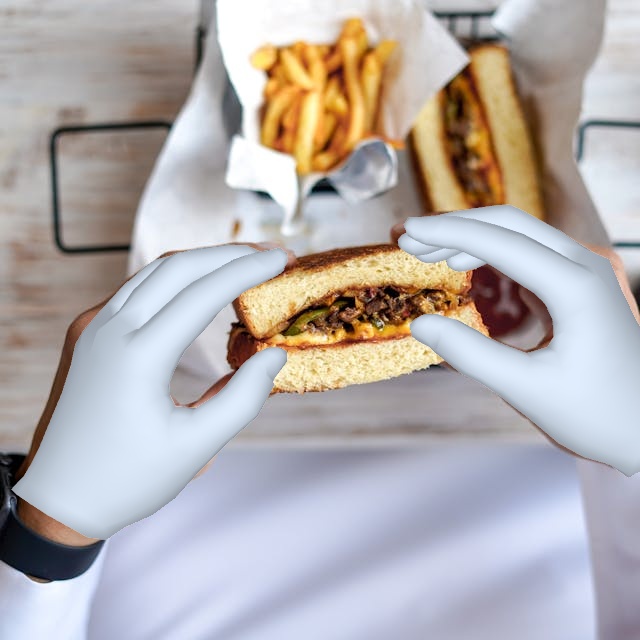}
        \end{subfigure} &
        \begin{subfigure}{0.14\textwidth}
            \includegraphics[width=\linewidth,height=\linewidth,keepaspectratio=false]{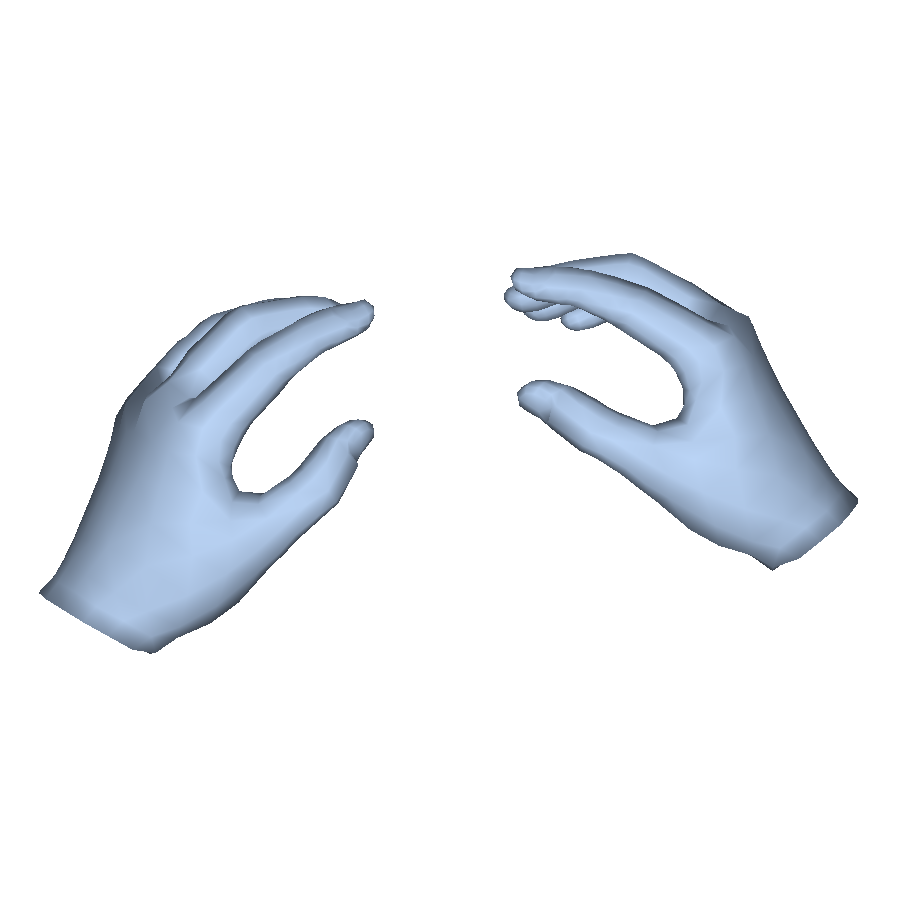}
        \end{subfigure} &
        \begin{subfigure}{0.14\textwidth}
            \includegraphics[width=\linewidth,height=\linewidth,keepaspectratio=false]{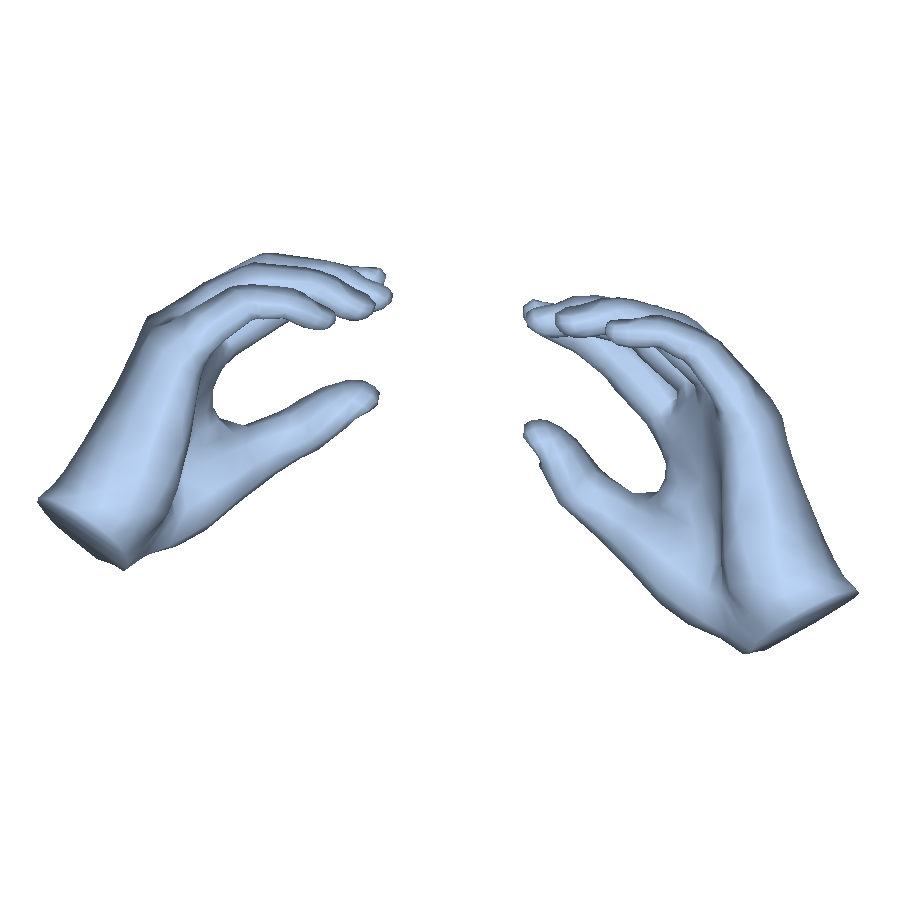}
        \end{subfigure} \\

        \begin{subfigure}{0.14\textwidth}
            
            \includegraphics[width=\linewidth,height=\linewidth,keepaspectratio=false]
            {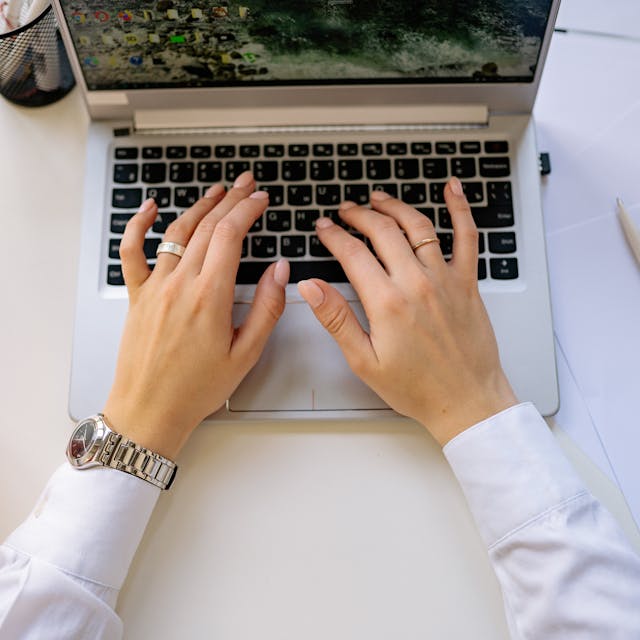}

        \end{subfigure} &
        \begin{subfigure}{0.14\textwidth}
            
            \includegraphics[width=\linewidth,height=\linewidth,keepaspectratio=false]{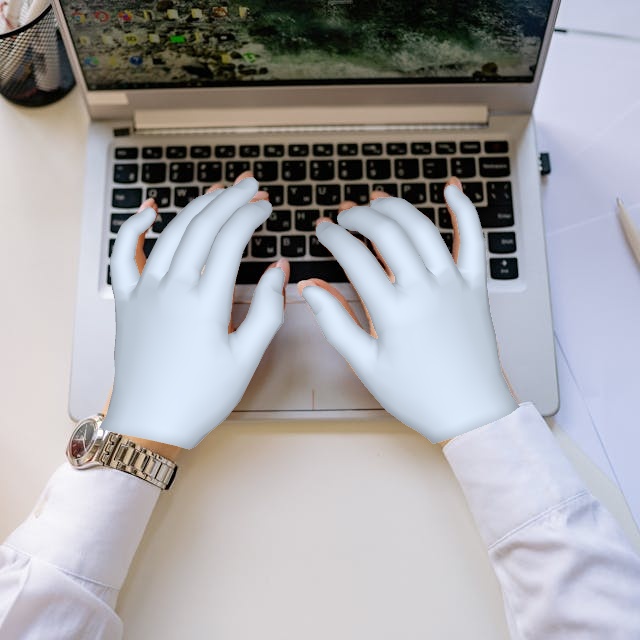}
        \end{subfigure} &
        \begin{subfigure}{0.14\textwidth}
            
            \includegraphics[width=\linewidth,height=\linewidth,keepaspectratio=false]{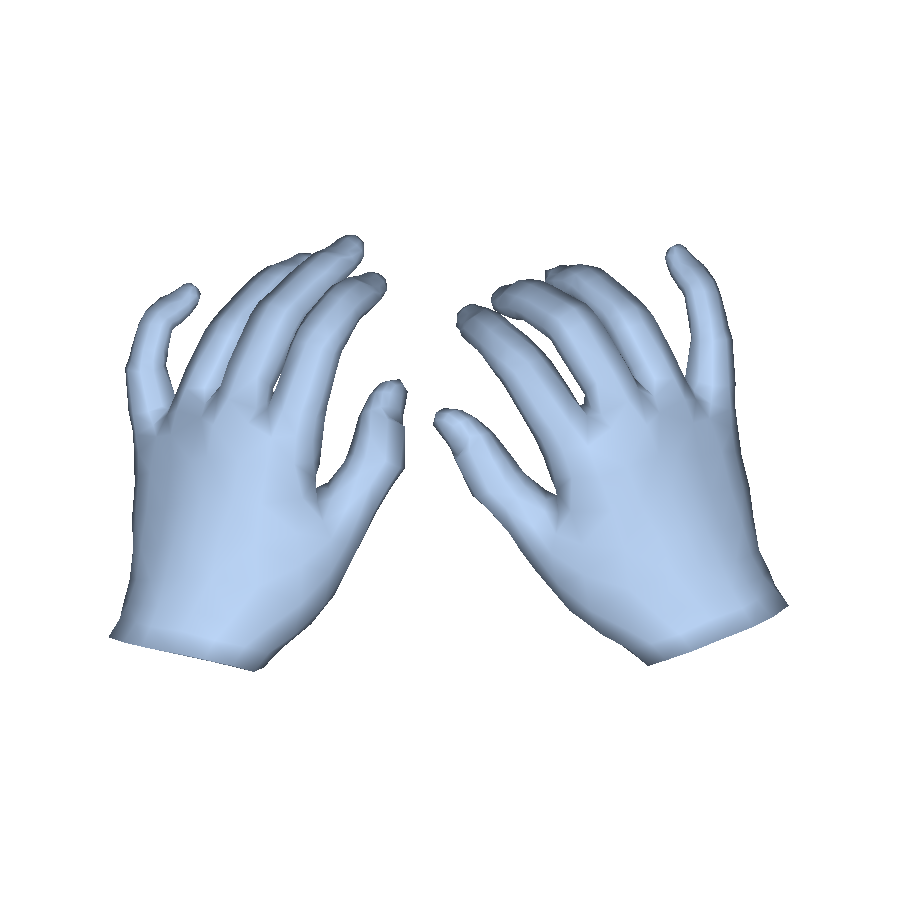}
        \end{subfigure} &
        \begin{subfigure}{0.14\textwidth}
            
            \includegraphics[width=\linewidth,height=\linewidth,keepaspectratio=false]{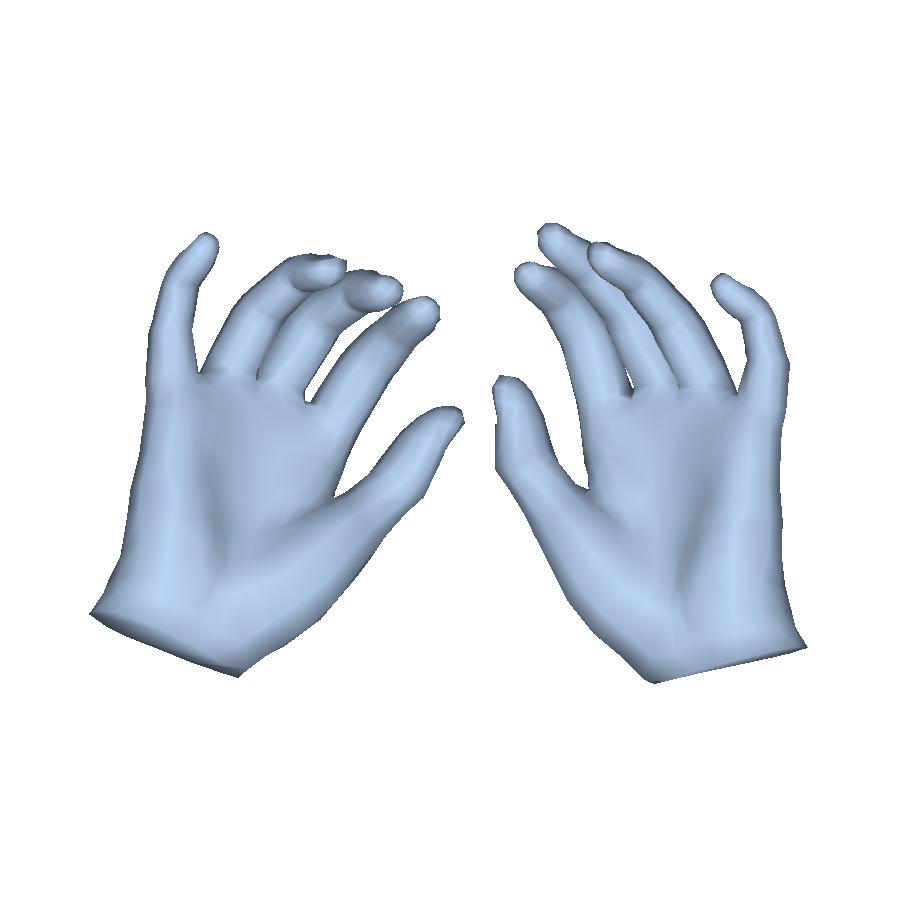}
        \end{subfigure} &
        \begin{subfigure}{0.14\textwidth}
            
            \includegraphics[width=\linewidth,height=\linewidth,keepaspectratio=false]{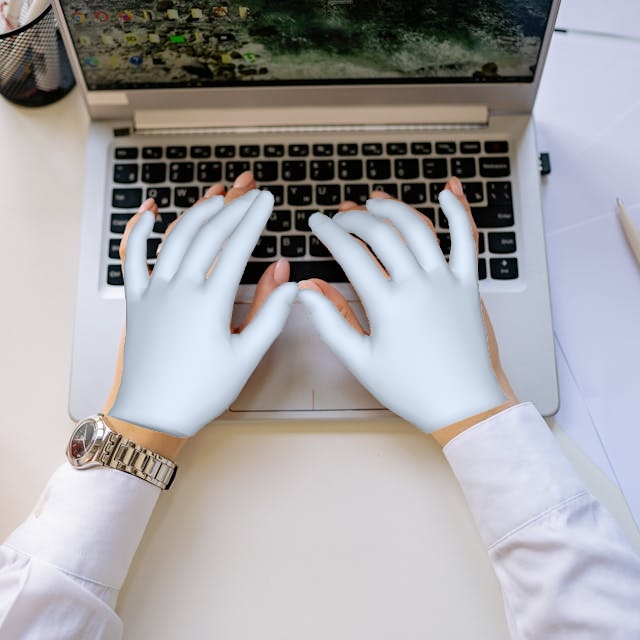}
        \end{subfigure} &
        \begin{subfigure}{0.14\textwidth}
            
            \includegraphics[width=\linewidth,height=\linewidth,keepaspectratio=false]{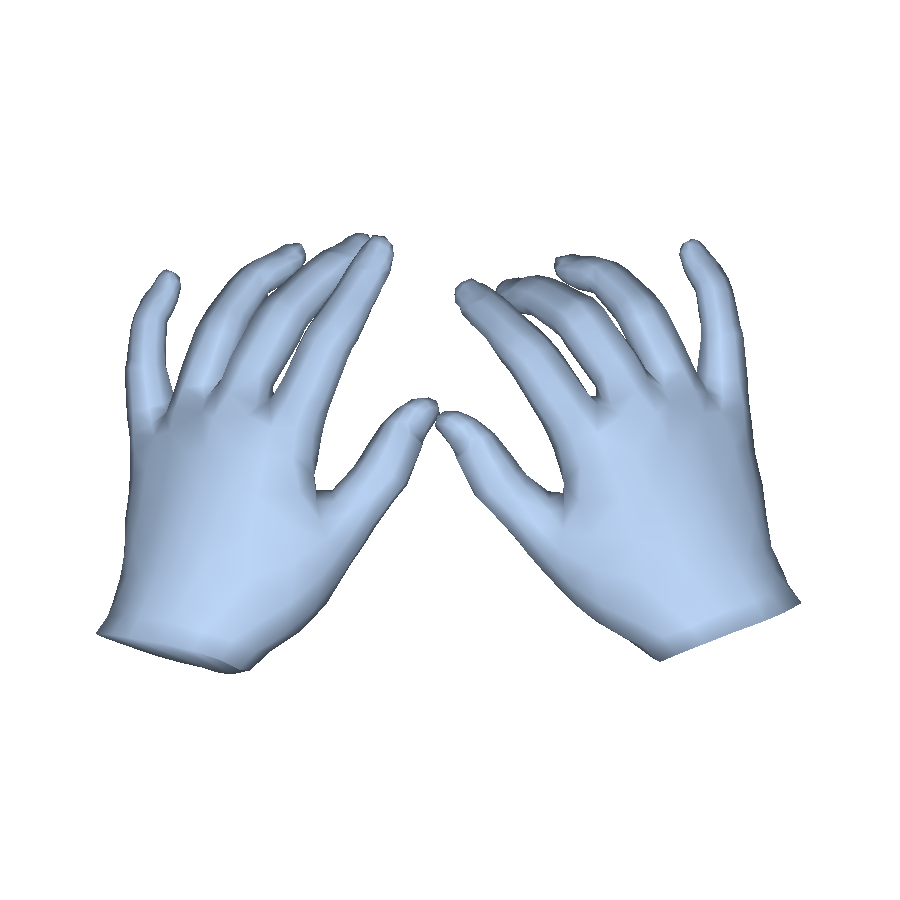}
        \end{subfigure} &
        \begin{subfigure}{0.14\textwidth}
            
            \includegraphics[width=\linewidth,height=\linewidth,keepaspectratio=false]{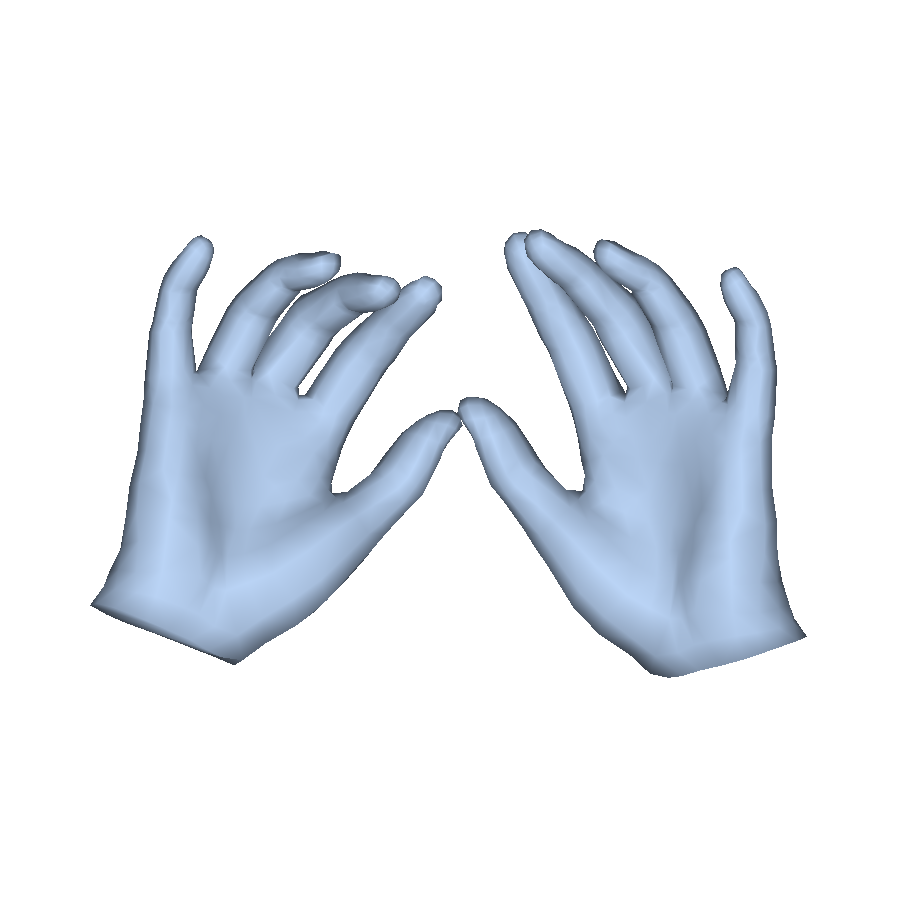}
        \end{subfigure} \\
        
    \end{tabular}
    }
    \caption{Qualitative results on images from the internet, the scenes represent hands interacting with the environment. 
    We compare our results of the best configuration (ConvNext-L with feature-level distillation) with HaMeR both in 2D and 3D.}
    \label{fig:qualitative-results}
\end{figure*}
\subsubsection{Output-Level Knowledge Distillation}

\begin{table}[t]
\setlength{\tabcolsep}{6pt} 
\centering
\caption{\textbf{Output-level Distillation:} Evaluating different backbones trained with KD loss applied on the outputs of the student and teacher models. $\lambda_{KD}$ refers to the weight of the KD loss.}
\resizebox{\columnwidth}{!}{%

\begin{tabular}{@{}lcccc@{}}
\toprule
\textbf{Backbone} &
  \bm{$\lambda_{KD}$} &
    \textbf{\begin{tabular}[c]{@{}c@{}}$\mathbf{J}_{err}$ $\downarrow$\end{tabular}} & 
    \textbf{\begin{tabular}[c]{@{}c@{}}$\mathbf{V}_{err}$ $\downarrow$\end{tabular}} & 
  \multicolumn{1}{c}{\textbf{\begin{tabular}[c]{@{}c@{}}F@5.0 / F@15.0 $\uparrow$\end{tabular}}} \\ \midrule
  
MobileNet-L & 0.3 & 9.5 & 9.6 & 0.525 / 0.964 \\
MobileNet-L & 0.5 & 9.3 & 9.4 & 0.528 / 0.964 \\
MobileNet-L & 0.8 & 9.3 & 9.4 & 0.525 / 0.964 \\ \midrule

MobileViT-S & 0.3 & 9.1 & 9.2 & 0.536 / 0.968 \\
MobileViT-S & 0.5 & 9.0 & 9.1 & 0.543 / 0.970 \\
MobileViT-S & 0.8 & 9.5 & 9.6 & 0.516 / 0.965 \\ \midrule

ResNet-50 & 0.3 & 9.0 & 9.2 & 0.547 / 0.969 \\
ResNet-50 & 0.5 & 8.9 & 9.0 & 0.562 / 0.968 \\
ResNet-50 & 0.8 & 9.0 & 9.1 & 0.556 / 0.970 \\ \midrule

ResNet-101 & 0.3 & 8.9 & 9.1 & 0.552 / 0.970 \\
ResNet-101 & 0.5 & 8.8 & 8.9 & 0.569 / 0.970 \\
ResNet-101 & 0.8 & 8.5 & 8.7 & 0.583 / 0.974 \\ \midrule

ConvNext-L & 0.3 & 8.3 & 8.5 & 0.588 / 0.976 \\
\rowcolor{lightred} ConvNext-L & 0.5 & 8.2 & 8.4 & 0.596 / 0.978 \\
ConvNext-L & 0.8 & 8.4 & 8.7 & 0.582 / 0.974 \\ \bottomrule

\end{tabular}%
}
\label{tab:output-level}
\end{table}

Next, we evaluate output-level distillation, where the objective is to allow the student network to learn from the predictions of the teacher network along with the ground-truth. 
The formulation of all the relevant loss terms is detailed in Section \ref{sec:kd}. 
All the results in this study are summarized in Table~\ref{tab:output-level}.
Applying teacher supervision on the outputs of the student model produces interesting results across different $\lambda_{KD}$ (KD weight) values. 
For MobileNet-L \cite{Howard2017MobileNetsEC} configurations, we observe no improvements in any of the metrics; instead, we observe a degradation of up to \textit{0.3mm} for PA-MPVPE, along with similar degradation for all other metrics. 
On the other hand, MobileViT-S \cite{Mehta2021MobileViTLG} configurations offer noticeable improvements at $\lambda_{KD} = 0.5$, with \textit{0.4mm} improvement in PA-MPVPE. 
This gain could indicate the compatibility between the similar architecture shared by both the ViT-Huge \cite{dosovitskiy2021imageworth16x16words} teacher backbone and the MobileVit-S student backbone \cite{Mehta2021MobileViTLG}.
The improvement, however, starts to degrade at $\lambda_{KD} = 0.8$. 
ResNet-50 \cite{He2015DeepRL} does not yield noticeable improvements with output-level distillation; the baseline without any distillation remains the best-performing configuration.
ResNet-101 \cite{He2015DeepRL} with a deeper CNN architecture suggests some improvements at $\lambda_{KD} = 0.8$ with approximately \textit{0.3mm} improvement in PA-MPVPE. 
The configuration with ConvNeXt-L \cite{Liu2022ACF} and $\lambda_{KD} = 0.5$ shows a small improvement by \textit{0.1mm} in PA-MPVPE.

\subsubsection{Feature-Level Knowledge Distillation}

In this section, we further explore the performance of the models with distillation applied on the feature-map level. 
More specifically, in these configurations, the backbone in the student network attempts to learn a similar feature representation as the larger teacher model. 
The evaluation results are presented in Table \ref{tab:feature-distillation-results}. 




We observe that MobileNet-L \cite{Howard2017MobileNetsEC}, MobileViT-S \cite{Mehta2021MobileViTLG}, ResNet-50, and ResNet-100 \cite{He2015DeepRL} all show no improvement with feature-level distillation, with some even performing worse than output-level distillation.
Interestingly, ConvNeXt-L~\cite{Liu2022ACF} performs best under feature-level distillation, with $\lambda_{KD}=0.8$ and $\gamma_{FD}=12$, it was able to achieve the best results across all the metrics, showing a \textit{0.2mm} improvement in PA-MPVPE compared to its baseline. 
The results for ConvNeXt suggest that feature-level distillation might be more suitable for relatively high-capacity student networks. 
A network like ConvNeXt with deep and modern architecture and rich hierarchical features might be better suited to learn the spatial and semantic cues in the teacher's feature maps.

\subsubsection{Combined Distillation: Feature + Output}

Finally, we evaluate the performance of the student networks trained with a combined distillation loss, where knowledge is transferred at both feature maps and output levels. 
This setting aims to leverage the information from both loss types and mimic the teacher more holistically. 
The quantitative results for this strategy are reported in Table \ref{tab:combined-knowledge-distillation}.


We observe that for most backbones, combined distillation achieves performance between that of individual output-level and feature-level strategies. 
The combination of both distillation methods (i.e., output-level and feature-level) generally does not surpass the individual methods. 
For instance, ConvNeXt-L configurations do not show the same improvements gained by feature-level distillation.
We can also observe this for ResNet-101, where combined distillation offers only a marginal increase in performance. 
The results from these experiments show that while it is possible to obtain some improvements over non-distilled models, specifically for relatively larger models, a combined distillation approach may dilute the stronger signal of either individual method.

\begin{table}[]
\setlength{\tabcolsep}{6pt}
\centering
\caption{\textbf{Feature-level Distillation:} Evaluating different backbones trained with KD loss applied on the feature maps produced by the backbones of student and teacher models. \textbf{$\lambda_{KD}$} refers to the weight of the total KD loss, while \textbf{$\gamma_{FD}$} refers to a scalar used to scale only the feature-distillation loss.}
\resizebox{1.0\columnwidth}{!}{%

\begin{tabular}{@{}lccccc@{}}
\toprule
\textbf{Backbone} &
  $\boldsymbol{\lambda_{KD}}$ &
  $\boldsymbol{\gamma_{FD}}$ &
    \textbf{\begin{tabular}[c]{@{}c@{}}$\mathbf{J}_{err}$ $\downarrow$\end{tabular}} & 
    \textbf{\begin{tabular}[c]{@{}c@{}}$\mathbf{V}_{err}$ $\downarrow$\end{tabular}} & 
  \multicolumn{1}{c}{\textbf{\begin{tabular}[c]{@{}c@{}}F@5.0 / F@15.0 $\uparrow$\end{tabular}}} \\ \midrule
  
MobileNet-L & 0.3 & 6 & 9.6 & 9.7 & 0.514 / 0.961 \\
MobileNet-L & 0.5 & 6 & 10.4 & 10.5 & 0.460 / 0.952 \\
MobileNet-L & 0.8 & 12 & 10.0 & 10.2 & 0.477 / 0.958 \\ \midrule

MobileViT-S & 0.3 & 6 & 9.5 & 9.4 & 0.525 / 0.965 \\
MobileViT-S & 0.5 & 6 & 9.8 & 9.9 & 0.491 / 0.961 \\
MobileViT-S & 0.8 & 12 & 9.7 & 9.7 & 0.509 / 0.959 \\ \midrule

ResNet-50 & 0.3 & 6 & 9.1 & 9.2 & 0.541 / 0.969 \\
ResNet-50 & 0.5 & 6 & 9.2 & 9.3 & 0.531 / 0.968 \\
ResNet-50 & 0.8 & 12 & 9.0 & 9.0 & 0.546 / 0.971 \\ \midrule

ResNet-101 & 0.3 & 6 & 9.0 & 9.1 & 0.547 / 0.970 \\
ResNet-101 & 0.5 & 6 & 8.9 & 9.0 & 0.546 / 0.970 \\
ResNet-101 & 0.8 & 12 & 9.0 & 9.0 & 0.546 / 0.971 \\ \midrule

ConvNext-L & 0.3 & 6 & 8.5 & 8.6 & 0.579 / 0.978 \\
ConvNext-L & 0.5 & 6 & 8.2 & 8.4 & 0.591 / 0.977 \\
\rowcolor{lightred} ConvNext-L & 0.8 & 12 & 8.1 & 8.3 & 0.599 / 0.979 \\ \bottomrule

\end{tabular}%
}
\label{tab:feature-distillation-results}
\end{table}

\begin{table}[]
\setlength{\tabcolsep}{6pt}
\centering
\caption{\textbf{Combined Distillation:} Evaluating different backbones trained with KD loss applied on the outputs and the feature maps produced by the backbones of student and teacher models. \textbf{$\lambda_{KD}$} refers to the weight of the total KD loss, while \textbf{$\gamma_{FD}$} refers to a scalar used to scale only the feature-distillation loss.}
\resizebox{\columnwidth}{!}{%

\begin{tabular}{@{}lccccc@{}}
\toprule
\textbf{Backbone} &
  \bm{$\lambda_{KD}$} &
  \bm{$\gamma_{FD}$} &
    \textbf{\begin{tabular}[c]{@{}c@{}}$\mathbf{J}_{err}$ $\downarrow$\end{tabular}} & 
    \textbf{\begin{tabular}[c]{@{}c@{}}$\mathbf{V}_{err}$ $\downarrow$\end{tabular}} & 
  \multicolumn{1}{c}{\textbf{\begin{tabular}[c]{@{}c@{}}F@5.0 / F@15.0 $\uparrow$\end{tabular}}} \\ \midrule
MobileNet-L & 0.3 & 6 & 9.6 & 9.7 & 0.509 / 0.960 \\
MobileNet-L & 0.5 & 6 & 9.8 & 9.9 & 0.499 / 0.961 \\
MobileNet-L & 0.8 & 12 & 10.3 & 10.4 & 0.472 / 0.951 \\ \midrule

MobileViT-S & 0.3 & 6 & 9.8 & 9.8 & 0.498 / 0.960 \\
MobileViT-S & 0.5 & 6 & 9.7 & 9.8 & 0.498 / 0.959 \\
MobileViT-S & 0.8 & 12 & 9.9 & 10.0 & 0.489 / 0.958 \\ \midrule

ResNet-50 & 0.3 & 6 & 9.1 & 9.2 & 0.540 / 0.969 \\
ResNet-50 & 0.5 & 6 & 9.2 & 9.2 & 0.538 / 0.969 \\
ResNet-50 & 0.8 & 12 & 9.2 & 9.3 & 0.533 / 0.968 \\ 
\midrule

ResNet-101 & 0.3 & 6 & 9.0 & 9.1 & 0.551 / 0.970 \\
ResNet-101 & 0.5 & 6 & 8.9 & 9.1 & 0.553 / 0.970 \\
ResNet-101 & 0.8 & 12 & 9.0 & 9.2 & 0.550 / 0.969 \\ \midrule

ConvNext-L & 0.3 & 6 & 8.6 & 8.8 & 0.573 / 0.975 \\
ConvNext-L & 0.5 & 6 & 8.5 & 8.7 & 0.581 / 0.976 \\
\rowcolor{lightred} ConvNext-L & 0.8 & 12 & 8.4 & 8.5 & 0.587 / 0.978 \\ \bottomrule

\end{tabular}%
}
\label{tab:combined-knowledge-distillation}
\end{table}

\begin{table}[]
\setlength{\tabcolsep}{1pt} 
\centering
\caption{Comparison with the state-of-the-art using our best network configuration. 
The selected model contains a ConvNeXt-L \cite{Liu2022ACF} as the backbone and was trained using feature-level distillation with $\lambda_{KD} = 0.8, \gamma_{FD} = 12$.}
\resizebox{\columnwidth}{!}{%

\begin{tabular}{@{}lccc@{}}
\toprule
\textbf{Method} & 
    \textbf{\begin{tabular}[c]{@{}c@{}}$\mathbf{J}_{err}$ $\downarrow$\end{tabular}} & 
    \textbf{\begin{tabular}[c]{@{}c@{}}$\mathbf{V}_{err}$ $\downarrow$\end{tabular}} & 
    \textbf{\begin{tabular}[c]{@{}c@{}}F@5.0 / @15.0 $\uparrow$\end{tabular}} \\ \midrule
Pose2Mesh \cite{choi2020pose2mesh} & 12.5 & 12.7 & 0.441 / 0.909 \\
THOR-Net \cite{Aboukhadra_2023_WACV} & 11.3 & 10.7 & - / - \\
I2L-MeshNet \cite{Moon2020I2LMeshNetIP} & 11.2 & 13.9 & 0.409 / 0.932 \\
ArtiBoost \cite{Li2021ArtiBoostBA} & 11.1 & 10.9 & 0.488 / 0.944  \\
METRO \cite{Lin2020EndtoEndHP} & 10.4 & 11.1 & 0.484 / 0.946 \\
I2UV-HandNet \cite{chen2021i2uv} & 9.9 & 9.5 & 0.956 / 0.803 \\
HandOccNet \cite{park2022handoccnet} & 9.1 & 8.8 & 0.564 / 0.968 \\
HaMeR \cite{Pavlakos_2024_CVPR} & 7.7 & 7.9 & 0.635 / 0.980 \\
Hamba \cite{dong2024hamba} & 7.5 & 7.7 & 0.648 / 0.982 \\
WiLoR \cite{potamias2025wilor} & 7.5 & 7.7 & 0.646 / 0.983 \\ \midrule
Ours & 8.1 & 8.3 & 0.599 / 0.979 \\ \bottomrule
\end{tabular}%
}
\label{tab:comparision-with-sota}
\end{table}

\subsection{Comparison with the State-of-the-art}

In Table \ref{tab:comparision-with-sota}, we compare our best-performing model with recent state-of-the-art methods for 3D hand reconstruction. 
Using a ConvNeXt-L \cite{Liu2022ACF} backbone trained using feature-level distillation ($\lambda_{KD} = 0.8$, $\gamma_{FD} = 12$), our approach outperforms strong baselines including HandOccNet \cite{park2022handoccnet}, METRO \cite{Lin2020EndtoEndHP}, ArtiBoost \cite{Li2021ArtiBoostBA} and I2L-MeshNet \cite{Moon2020I2LMeshNetIP}. 
Our method achieves comparable performance to the recent methods, while also improving upon the inference time as discussed in Section \ref{sec:baselines}.
In Figure \ref{fig:qualitative-results}, we present qualitative results for our best-performing network configuration (i.e., ConvNeXt-L with feature-level Distillation) on challenging images collected from the internet. 
We also present results from HaMeR (teacher network) to demonstrate the difference in reconstruction accuracy. 
Although HaMeR’s qualitative results align more closely with the hands in the images, KD-trained ConvNeXt achieves similarly strong performance with only minor inaccuracies.





\vspace{-3mm}
\section{CONCLUSION}

In this work, we systematically investigated the use of smaller and lighter networks combined with Knowledge Distillation strategies—specifically output-level, feature-level, and their combination—to accelerate 3D hand mesh reconstruction.
Our experiments revealed key insights into the relationship between model size, architectural complexity, and the effectiveness of teacher-student supervision. 
We replaced HaMeR’s original ViT-H backbone with more efficient alternatives such as MobileViT and ConvNeXt, and evaluated the impact of different distillation strategies. 
Our results show that output-level distillation consistently yields strong performance across student models, while feature-level distillation provides benefits primarily for higher-capacity networks. 
Overall, this work demonstrates the potential of KD to enable fast and accurate hand reconstruction suitable for real-time applications on resource-constrained devices.
Our approach offers a practical alternative to large transformer-based models without a significant sacrifice in accuracy.



Future research can explore and investigate more expressive ways of enhancing feature-level distillation, such as using attention-based distillation, spatial or channel-wise distillation, or relational knowledge distillation. 
These approaches may better help the student networks to learn the complex feature representation of the teacher network.

\textbf{Acknowledgments}
This work was partially funded by the European Union under Horizon Europe under the project \textit{SHARESPACE} (101092889) and by the Federal Ministry of Education and Research Germany under the project \textit{COPPER} (16IW24009).




\FloatBarrier

\bibliographystyle{apalike}
{\small
\bibliography{FastHaMeR/references}}



\end{document}